\documentclass[11pt]{article}
\usepackage{microtype}

\usepackage[final]{acl}

\usepackage{times}
\usepackage{latexsym}
\usepackage{booktabs}
\usepackage{tikz}
\usepackage{pgfplots}
\usepackage{siunitx}
\usepackage{tabularx}
\usepackage{longtable}
\usepackage{makecell}
\usetikzlibrary{positioning}
\usepackage{forest}
\usepgfplotslibrary{groupplots}
\pgfplotsset{compat=1.18}
\usepackage{adjustbox}
\usetikzlibrary{calc,arrows.meta}
\usepackage{xcolor}
\usepackage[T1]{fontenc}
\usepackage{array}
\usepackage{listings}
\usepackage[most]{tcolorbox}
\usepackage{placeins}
\usepackage{stfloats}
\usepackage{listings}
\usepackage[most]{tcolorbox}
\usepackage{capt-of}

\usepackage{soul}

\definecolor{LightBlue}{rgb}{0.68, 0.85, 0.9}

\forestset{
  tagtree/.style={
    for tree={
      grow'=east,                 % <-- horizontal (left to right)
      parent anchor=east,
      child anchor=west,
      edge={->},
      draw,
      rounded corners=2pt,
      font=\ttfamily\scriptsize,
      inner sep=2.5pt,
      l sep=10pt,                 % horizontal level spacing
      s sep=7pt,                  % vertical sibling spacing
      align=left
    }
  }
}

\definecolor{csBlue}{HTML}{377EB8}  
\definecolor{csOrange}{HTML}{E69F00} 
\definecolor{csGreen}{HTML}{4DAF4A} 
\definecolor{csRed}{HTML}{E41A1C}
\definecolor{csPurple}{HTML}{984EA3} 
\definecolor{csGray}{HTML}{666666}

\pgfplotsset{
  bm25/.style={
    semithick,
    color=csGray,
    densely dashed,
    mark=*,
    mark size=1.2pt,
    mark options={solid, fill=white},
    /tikz/line cap=round,
    /tikz/line join=round
  },
  e5/.style={
    semithick,
    color=csBlue,
    mark=square*,
    mark size=1.2pt,
    mark options={solid, fill=white},
    /tikz/line cap=round,
    /tikz/line join=round
  },
  mbbase/.style={
    semithick,
    color=csOrange,
    densely dotted,
    mark=triangle*,
    mark size=1.4pt,
    mark options={solid, fill=white},
    /tikz/line cap=round,
    /tikz/line join=round
  },
  ftflat/.style={
    semithick,
    color=csGreen,
    dashdotted,
    mark=diamond*,
    mark size=1.3pt,
    mark options={solid, fill=white},
    /tikz/line cap=round,
    /tikz/line join=round
  },
  ftstruct/.style={
    semithick,
    color=csRed,
    mark=star,
    mark size=1.0pt,
    mark options={solid, fill=white},
    /tikz/line cap=round,
    /tikz/line join=round
  },
  random/.style={
    black!70,
    densely dashed,
    mark=none,
    line width=0.8pt
  }
}

\newcommand{\sectag}[1]{\textcolor{blue!70!black}{\texttt{#1}}}
\newcommand{\roletag}[1]{\textcolor{teal!70!black}{\texttt{#1}}}
\newcommand{\statustag}[1]{\textcolor{red!70!black}{\texttt{#1}}}
\newcommand{\masktag}[1]{\textcolor{violet!80!black}{\texttt{#1}}}

\definecolor{ConclusionRed}{HTML}{B84A4A}
\definecolor{AnalysisBlue}{HTML}{3F6FA6}
\definecolor{RuleGreen}{HTML}{4F8A63}
\definecolor{ContextGold}{HTML}{B18435}
\definecolor{MotivationNavy}{HTML}{24364B}

\newtcblisting{promptbox}[2][]{%
    enhanced,
    listing only,
    width=\linewidth,
    colback=black!2,
    colframe=black!35,
    boxrule=0.4pt,
    arc=1mm,
    left=1.5mm,
    right=1.5mm,
    top=1.5mm,
    bottom=1.5mm,
    before skip=6pt,
    after skip=6pt,
    title={#2},
    fonttitle=\bfseries\small,
    coltitle=black,
    colbacktitle=black!7,
    listing options={
      language=Python,
      basicstyle=\ttfamily\footnotesize,
      commentstyle=\upshape,
      columns=fullflexible,
      keepspaces=true,
      showstringspaces=false,
      breaklines=true,
      breakatwhitespace=false,
      tabsize=4,
      numbers=none
  },
    #1
}

\usepackage[T1]{fontenc}
\usepackage[utf8]{inputenc}

\usepackage{microtype}

\usepackage{inconsolata}

\usepackage{graphicx}

\title{Mining Legal Arguments in U.S. Corporate Case Law}

\author{
  \textbf{Luis Brena\textsuperscript{1}},
  \textbf{William Jurayj\textsuperscript{1}},
  \textbf{Gregory Deyesu\textsuperscript{2}},
  \textbf{Zaid Al-Huneidi\textsuperscript{2}},
  \\
  \textbf{Andrew Blair-Stanek\textsuperscript{1,2}},
  \textbf{Benjamin Van Durme\textsuperscript{1}}
  \\
  \\
  \textsuperscript{1}Johns Hopkins University \\
  \textsuperscript{2}University of Maryland School of Law
  \\
  {\small\texttt{\{lbrenap1,wjurayj1,vandurme\}@jhu.edu}} \\
  {\small\texttt{\{gdeyesu,zaidal-huneidi\}@umaryland.edu}} \\
  {\small\texttt{ablair-stanek@law.umaryland.edu}}
}

\begin{document}
\maketitle
\begin{abstract}

Legal argument mining supports passage classification, retrieval, and argument completion. This work introduces an expert-annotated dataset of 42 U.S. federal tax opinions on corporate reorganizations under I.R.C. \S368. To our knowledge, it is the first expert-annotated, tree-structured argument corpus for this domain. Explicit spans receive one of five functional labels: \textit{Rule}, \textit{Analysis}, \textit{Conclusion}, \textit{Background Facts}, and \textit{Procedural History}. Rule, Analysis, and Conclusion spans can be linked into directed support trees, while Background Facts and Procedural History serve a contextual function. The corpus provides span-based, sentence-based, flat, and tree-structured representations. Agreement analysis shows that functional node labels are more reliable than directed support edges and implicit intermediate conclusions. Directed-path agreement is stronger than direct-edge agreement, which indicates that broad reachability is more stable than exact local decomposition. Classification experiments show that functional labels are learnable under case-disjoint evaluation. Retrieval experiments show that supervised fine-tuning improves within-case retrieval. However, cross-case generalization remains weak. The dataset supports legal passage classification and provides a conservative benchmark for structured argument mining in U.S. federal tax case law.

% \st{The proposed scheme demonstrates recursion, deductive closure, and structural logic mapping properties. Positive results were obtained in both classification and retrieval experiments. 

\end{abstract}

\section{Introduction}

Legal argument mining has produced resources across jurisdictions and tasks, including clause-level mining in European Court of Human Rights (ECHR) decisions \citep{poudyal-etal-2020-echr,habernal-etal-2024-mining}, argument-scheme annotation in Court of Justice of the European Union (CJEU) fiscal state-aid decisions \citep{grundler-etal-2022-detecting,santin-2023-relation-prediction}, functional and rhetorical-role labeling in U.S. judicial opinions \citep{walker-etal-2017-semantic,savelka-ashley-2018-segmenting,belfathi-etal-2026-coupling}, U.S. civil-procedure argument reasoning \citep{bongard-etal-2022-legal}, and U.S. case-law retrieval \citep{hou-etal-2025-clerc}. Together, these resources show that legal reasoning can be modeled at several granularities, from clause and sentence roles to premise--conclusion links, legal-reasoning tasks, and retrieval targets. Yet U.S. case law remains underrepresented in legal NLP, and existing resources do not provide expert-adjudicated, document-level support trees for this domain.

This study examines U.S. federal tax opinions on corporate reorganizations under I.R.C.\ \S368. The corpus includes opinions involving \S368(a)(1)(A), (B), (C), (D), and (F), covering statutory mergers and consolidations, stock acquisitions, asset acquisitions, certain asset transfers, and changes in corporate identity, form, or place of organization. Recapitalizations under \S368(a)(1)(E) and bankruptcy reorganizations under \S368(a)(1)(G) are excluded to limit statutory variety \citep{usc_368}. This narrow domain is computationally useful because courts must connect statutory categories, continuity-of-business-enterprise and continuity-of-interest requirements, transaction-specific facts, and intermediate legal conclusions \citep{cfr_1368_1}. Because U.S. courts operate in a precedent-based system, these opinions also support analysis of how statutory rules, case facts, and prior authorities interact in written legal reasoning \citep{valvoda2021precedent}. The domain therefore allows annotations beyond merely labeling isolated sentences, providing a testbed for legal NLP systems that model how rules and facts support downstream conclusions.

\begin{figure*}
    \centering
    \includegraphics[width=1\linewidth]{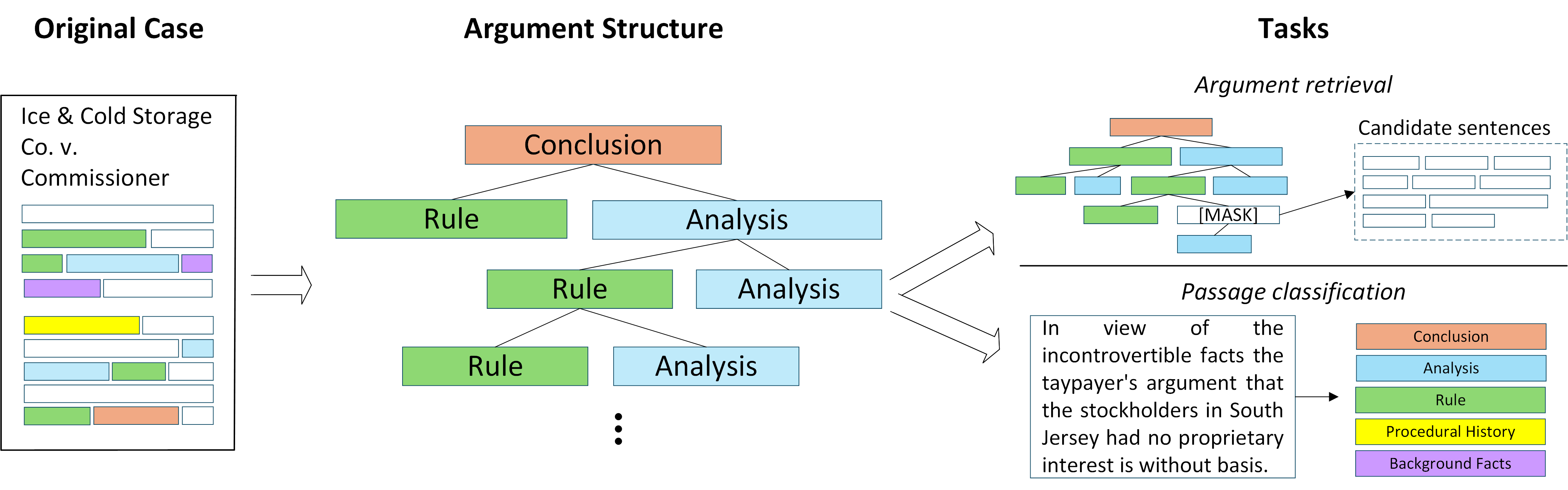}
    \caption{Overview of the annotation method and the associated tasks. In the annotation process, annotators selected spans of text without length constraints and labeled them as \textit{Conclusion}, \textit{Rule}, \textit{Analysis}, \textit{Background Facts}, or \textit{Procedural History}. Annotators connected these spans based on their roles within the argument structure. The annotation enabled two primary tasks: passage classification and masked-slot argument retrieval.}
    \label{fig:motivation-scheme}
\end{figure*}

\paragraph{Contributions.}
This paper presents three primary contributions:
\begin{enumerate}
    \item We release an expert-annotated corpus comprising 42 U.S. federal tax opinions on corporate reorganizations under I.R.C. \S368.\footnote{\url{https://huggingface.co/datasets/lbrenap1/mining-legal-arguments-us-corporate-case-law}}
    \item We provide span-based, sentence-based, flat, and tree-structured representations of functional legal-role annotations, alongside inter-annotator statistics that highlight reliability limits of the tree layer.
    \item We evaluate passage classification and argument-completion retrieval using case-disjoint splits, demonstrating that functional labels are learnable, whereas structure-aware retrieval exhibits limited generalization across cases (Figure~\ref{fig:motivation-scheme}).
\end{enumerate}

\section{Related Work}

Argument mining is commonly defined as the automatic identification and extraction of inference and reasoning structures in natural-language arguments \citep{lawrence-reed-2019-argument}. In legal NLP, this has led to work on corpora with explicit argument units and relations, rhetorical-role and functional segmentation, legal-reasoning benchmarks, and retrieval resources (Appendix~\ref{sec:related_work}).

\subsection{Legal argument-structure corpora}

Prior work has annotated legal argument structures across jurisdictions. \citet{yamada-etal-2019-corpus} developed an annotation scheme for Japanese judgments aimed at structure-based summarization. \citet{poudyal-etal-2020-echr} released an ECHR corpus with clause-level premise, conclusion, and non-argument labels, plus relations among argumentative clauses. The Demosthenes project is especially related: \citet{grundler-etal-2022-detecting} annotated CJEU fiscal state-aid decisions for argumentative elements, element types, and argument schemes, and \citet{santin-2023-relation-prediction} extended this work to argument-structure prediction. \citet{habernal-etal-2024-mining} further expanded ECHR argument mining with a span-level actor--argument-type scheme grounded in legal argumentation theory and ECHR practice.

Other jurisdictions provide additional models. Annotation guidelines for Chinese judicial decisions formalize proposition types and relations such as support, attack, joint, match, and identity \citep{chen-etal-2026-guidelines}. In North American work, \citet{walker-etal-2017-semantic} annotated U.S. veterans' claims decisions with sentence roles and propositional connective types, while \citet{xu-etal-2020-using,xu-ashley-2022-multigranularity} studied Issue--Reason--Conclusion structures and granularity choices for argument mining and summarization. \citet{jurayj2026language} take a different approach in U.S. Tax Law, using language models instead to parse rules and fact into digestible formalizations that offload argumentation to a symbolic prover \cite{wielemaker:2011:tplp}. Together, these studies show that legal argument units and relations have been used for related tasks, but not for expert-adjudicated support trees in U.S. federal tax reorganization opinions. 

\subsection{Rhetorical-role and functional segmentation}

A related line of work labels the discourse function of legal text without explicit support relations. \citet{bhattacharya-etal-2019-identification} introduced sentence-level rhetorical-role labeling for Indian Supreme Court judgments, and later Indian legal-document resources expanded this approach with larger corpora and more detailed role classes \citep{kalamkar-etal-2022-corpus,malik-etal-2022-semantic,nigam-etal-2025-legalseg}. In U.S. case law, \citet{savelka-ashley-2018-segmenting} segmented trade-secret and cyber-crime opinions into consecutive, non-overlapping parts using seven functional and issue-specific labels, with a CRF-based system identifying part boundaries rather than arbitrary argumentative spans or support links.

Rhetorical-role labeling has also been extended to new jurisdictions and models. \citet{csanyi-etal-2025-rrl} developed a sentence-level rhetorical-role classifier for Hungarian judicial decisions, while \citet{bambroo-etal-2025-marro} proposed MARRO, a multi-task, multi-headed-attention model evaluated on Indian and U.K. Supreme Court datasets. \citet{belfathi-etal-2026-coupling} introduced SCOTUS-Law, a U.S. Supreme Court corpus annotated at three levels of granularity for rhetorical sentence function. These resources are important precursors for labels such as facts, rules, analysis, and conclusions, but they generally do not encode document-level support-tree structure.

\subsection{IRAC-style legal reasoning benchmarks}

The Issue, Rule, Analysis, Conclusion (IRAC) reasoning framework is increasingly prevalent in legal NLP benchmarks and prompting methods. \citet{guha-etal-2023-legalbench} introduced LegalBench, a broad legal-reasoning benchmark developed with legal professionals. More specialized benchmarks use IRAC or IRAC-aligned structures for scenario analysis, patent decisions, and tax-penalty questions \citep{Kang2024BridgingLA,jang-etal-2025-pilot,choi-etal-2026-taxation}. Related prompting work shows that IRAC-derived prompts can improve legal entailment \citep{yu-etal-2022-legal-prompting}, and Chain of Logic uses an IRAC-inspired method for rule-based reasoning \citep{servantez-etal-2024-chain}. \citet{bongard-etal-2022-legal} introduced a U.S. civil-procedure legal-argument reasoning task with case introductions, questions, proposed arguments, and explanatory analyses. This literature supports the use of legal-reasoning categories, while also highlighting the lack of support-tree annotations for U.S. federal tax judicial opinions.

\subsection{Legal retrieval}

Legal retrieval research studies how systems identify authorities or passages relevant to legal analysis. CLERC is especially relevant: it is a U.S. case-law retrieval and retrieval-augmented generation corpus for retrieving citations that support specific legal analyses and for generating analysis from retrieved authorities \citep{hou2024gaps, hou-etal-2025-clerc}. By contrast, our retrieval task focuses on within-case argument completion and global multi-case passage retrieval using sentences derived from annotated support trees, making it narrower and more directly tied to argument structure.

\section{Corpus Creation}

\begin{table}[t]

\centering
\footnotesize
\setlength{\tabcolsep}{3pt}
\renewcommand{\arraystretch}{1.04}

\begin{tabular*}{\columnwidth}{@{\extracolsep{\fill}}lcccc@{}}
\toprule
& \multicolumn{2}{c}{$\alpha_u$} & \multicolumn{2}{c}{F1} \\
\cmidrule(lr){2-3}\cmidrule(l){4-5}
Label & Span & Sentence & Edit & Semantic \\
\midrule
Analysis           & 0.57 & 0.57 & 0.46 & 0.56 \\
Background facts   & 0.88 & 0.88 & 0.70 & 0.79 \\
Conclusion         & 0.72 & 0.72 & 0.70 & 0.85 \\
Procedural history & 0.45 & 0.46 & 0.47 & 0.59 \\
Rule               & 0.61 & 0.61 & 0.51 & 0.64 \\
\midrule
\textbf{Macro avg.} & \textbf{0.65} & \textbf{0.65} & \textbf{0.57} & \textbf{0.69} \\
\bottomrule
\end{tabular*}

\caption{Inter-annotator agreement on explicit functional-node labels for the 10 double-annotated cases. Span-level $\alpha_u$ is computed over annotated text spans, while sentence-level $\alpha_u$ projects span annotations to sentence units. Edit and Semantic denote F1 under edit-distance-based and semantic span pairing.}
\label{tab:iaa-explicit}
\end{table}
\begin{table}[t]
\centering
\scriptsize
\setlength{\tabcolsep}{3pt}
\renewcommand{\arraystretch}{1.05}
\begin{tabular*}{\columnwidth}{@{\extracolsep{\fill}}llrrrrr@{}}
\toprule
\multicolumn{7}{@{}l}{\textbf{(a) Implicit Nodes Contingency Counts}} \\
\midrule
Pairing & Label match & Ctx & YY & YN & NY & NN \\
\midrule
Edit & Blind & 64 & 2 & 7 & 14 & 41 \\
Semantic & Blind & 62 & 2 & 7 & 14 & 39 \\
Edit & Same label & 63 & 2 & 7 & 14 & 40 \\
Semantic & Same label & 64 & 2 & 7 & 15 & 40 \\
\bottomrule
\end{tabular*}

\vspace{0.6em}

\begin{tabular*}{\columnwidth}{@{\extracolsep{\fill}}llrrrrcc@{}}
\toprule
\multicolumn{8}{@{}l}{\textbf{(b) Implicit Nodes Agreement Summary}} \\
\midrule
Pairing & Label match & $P_o$ & $P_+$ & $P_-$ & $\kappa$ & $R_{A1/A2}$ & $E_{A1/A2}$ \\
\midrule
Edit & Blind & 0.67 & 0.16 & 0.80 & -0.02 & 0.14 / 0.25 & 9 / 24 \\
Semantic & Blind & 0.66 & 0.16 & 0.79 & -0.03 & 0.15 / 0.26 & 9 / 24 \\
Edit & Same label & 0.67 & 0.16 & 0.79 & -0.03 & 0.14 / 0.25 & 9 / 24 \\
Semantic & Same label & 0.66 & 0.15 & 0.78 & -0.04 & 0.14 / 0.27 & 9 / 26 \\
\bottomrule
\end{tabular*}
\caption{Agreement on implicit intermediate-conclusion (IC) insertion. Ctx is the number of evaluable comparison contexts after explicit spans were aligned. Pairing methods are edit distance (Edit) and semantic similarity (Semantic). Label match indicates whether explicit nodes could be paired across classes (Blind) or only within the same class (Same label). YY, YN, NY, and NN indicate whether annotator 1 and annotator 2 inserted an implicit IC in the same context. $P_o$ is observed agreement; $P_+$ and $P_-$ are positive and negative agreement for insertion vs. non-insertion; $R_{A1/A2}$ gives per-annotator insertion rates; $E_{A1/A2}$ gives the number of inserted implicit nodes that could be mapped to an evaluable context for agreement scoring.}

\label{tab:iaa-ic-contingency}
\end{table}

We collected 42 U.S. corporate reorganization cases, each between 1k and 10k words, focusing on I.R.C. \S368(a)(1)(A), (B), (C), (D), and (F), while excluding \S368(a)(1)(E) and (G) to limit statutory variety. The opinions range in citation year from 1935 to 1987; Appendix~\ref{sec:data_statement} reports the distribution by court, year, and main topics.

Two law students with backgrounds in tax and corporate law annotated all 42 documents. Before full annotation, they completed three calibration cases and six hours of guideline training, followed by iterative review and continued consultation of the guidelines and platform. Ten documents were independently double-annotated for inter-annotator agreement, and final annotations were adjudicated by a professor of law with expertise in tax law research and teaching. The annotators and adjudicator are listed as authors. Annotation was conducted in a customized version of Label Studio \citep{labelstudio} supporting span labeling, directed links, and tree visualization. Appendix~\ref{sec:dataset_statistics} summarizes corpus scale, label distributions, and tree-level structural properties.

\subsection{Annotation procedure}
Annotators selected free spans expressing atomic units of reasoning, assigned each a functional-node label, and added directed support links to form argument trees following a syllogistic pattern \citep{GardnerBartholomew2020}. Each linked step represents a local inference supporting a downstream claim. When a premise was implicit, annotators could mark enthymemes and, when needed, insert implicit intermediate conclusions to preserve structural consistency. Appendix~\ref{sec:annotation_workflow} illustrates the workflow.

\subsection{Argumentation scheme}
\paragraph{Labels.}
The scheme uses five labels.
\begin{itemize}
    \item \textit{Rule} marks generally applicable statements, including legal rules, tests, and abstract criteria.
    \item \textit{Analysis} marks case-specific reasoning applying rules to facts and often functioning as intermediate conclusions.
    \item \textit{Conclusion} marks the final outcome of an argument tree.
    \item \textit{Background Facts} marks case facts and transaction details that explain the factual setting but do not contribute directly to argumentative reasoning, such as corporate structure, stock ownership, asset transfers, trust arrangements, or business operations.
    \item \textit{Procedural History} marks litigation posture and procedural events, such as refund claims, deficiency notices, appeals, remands, and the court whose decision is under review.
\end{itemize}

\paragraph{Relations and constraints.}
Spans can be linked by directed support relations to form trees. \textit{Rule} and \textit{Analysis} spans belong to the argumentative structure and must have a directed path to a terminal \textit{Conclusion}. \textit{Background Facts} and \textit{Procedural History} may be annotated for future use but remain disconnected from the argument tree. \textit{Conclusion} spans are terminal nodes and cannot support other conclusions.

\paragraph{Intermediate implicit conclusions and enthymemes.}
Enthymemes are arguments that depend on one or more implicit premises \citep{feng-hirst-2011-classifying}. In our syllogistic scheme, an enthymeme is an abbreviated syllogism in which an unstated \textit{Analysis} or \textit{Rule} must be inferred from shared background knowledge, often through a causal relation. Annotators were instructed to use this label only when strictly necessary and when the missing component could not be found explicitly. Appendix~\ref{sec:annotation_structure} shows a condensed example.

\subsection{Dataset curation}

\paragraph{Adjudication.}
For the ten double-annotated cases, the adjudicator reviewed complete annotation files, including spans, labels, implicit nodes, and support links, and selected the version that best followed the guidelines. Adjudication therefore produced one final annotation file per case rather than discarding cases with disagreement or merging spans and edges across annotators. Disagreements were used for reliability analysis, while the released dataset reflects the adjudicated case-level files.

\paragraph{Span-to-sentence mapping.}
We generated a sentence-level version of the dataset for retrieval experiments and as an alternative annotation format. Character-interval overlap linked each sentence to any overlapping span node, with the sentence inheriting the node attributes and the passage text set to the full sentence. When a sentence overlapped multiple annotations, it was linked to all overlapping nodes for positive lookup, but the corpus record retained one original annotation ID and one label. Sentence identity was determined by maximum overlap, with ties resolved by earlier span start, earlier span end, and then lexicographic node ID. Before mapping, sentence-final periods and spaces were removed, and abbreviations, section names, and other non-sentence elements were incorporated into complete sentences.

\subsection{Inter-Annotator agreement}
Agreement was evaluated on ten independently double-annotated cases across four layers: explicit functional-node labels, implicit intermediate-conclusion insertion, directed support edges, and directed-path reachability.

\subsubsection{Functional-node label agreement}
Because annotators freely chose span boundaries, labels, and graph structure, there was no predefined set of shared spans. We therefore used two complementary label-agreement measures. First, Krippendorff's unitized $\alpha_u$ was computed separately for each label over character offsets \citep{krippendorff1995reliability}. Each label was treated as a binary segmentation task, with characters inside spans of that label marked positive and all others as background; the coincidence matrix was weighted by segment length.

Second, span-level soft-F1 measured sensitivity to boundary variation \citep{johansson-moschitti-2010-syntactic,tammewar-etal-2020-annotation}. Explicit spans were aligned using maximum-weight bipartite matching. The edit-distance condition used the Yujian-Bo edit-distance metric \citep{li-liu-2007-normalized-levenshtein}, while the semantic condition used cosine similarity between Cohere Embed v4 span embeddings with 512 output dimensions \citep{cohere-embed-v4-2025}. The Hungarian algorithm selected the one-to-one alignment with maximum total similarity \citep{kuhn1955hungarian}. Soft-F1 complements, but does not replace, $\alpha_u$: it scores matched span pairs after alignment, whereas $\alpha_u$ measures segmentation agreement over the full character sequence.

\paragraph{Results.}
Functional-node label agreement was the strongest annotation layer. In the span view, agreement was high for \textit{Background Facts} ($\alpha_u=0.879$) and \textit{Conclusion} ($\alpha_u=0.724$), and moderate for \textit{Rule} ($\alpha_u=0.613$), \textit{Analysis} ($\alpha_u=0.567$), and \textit{Procedural History} ($\alpha_u=0.454$). The sentence view produced nearly identical $\alpha_u$ values, showing that span-to-sentence conversion had little effect on unitized agreement. Soft-F1 was higher in the sentence view for \textit{Analysis}, \textit{Background Facts}, and \textit{Rule}, indicating that some disagreement reflects span-boundary variation rather than label choice alone (Table~\ref{tab:iaa-explicit}).

\subsubsection{Implicit nodes and directed-edge agreement}
Implicit-node and directed-edge agreement required pairing explicit nodes across annotators before comparing structural decisions. One-to-one regimes used maximum-weight bipartite matching based on either edit-distance or semantic similarity. Implicit-node insertion agreement is reported only for these one-to-one regimes. Directed-edge and directed-path agreement also use a relaxed offset-overlap regime, in which explicit spans were connected when their overlap covered at least a fixed fraction of the shorter span; connected components of these links formed non-overlapping comparison groups. Both label-blind and same-label pairing are reported to test sensitivity to label disagreement and span-granularity differences.

For structural scoring, platform relations were normalized to a parent-to-child direction. Directed-edge analysis treated each ordered pair of paired units as a binary decision indicating whether a directed support edge existed. The all-pairs context evaluates every ordered source-target pair and includes true negatives, while the edge-union context evaluates only pairs where at least one annotator proposed an edge. Because edge-union contains no true negatives, $P_-$, $\bar{P}_{\pm}$, and $\kappa$ are not reported there; $\kappa$ should be interpreted only for all-pairs results.

Implicit-node agreement was measured as insertion agreement. For each matched explicit parent that served as a structured parent in at least one annotation, we recorded whether each annotator inserted at least one implicit intermediate conclusion under that parent. This measures agreement on the decision to insert an implicit reasoning step, not agreement on the exact content of the implicit node.

\paragraph{Results.}
Implicit-node insertion showed high observed agreement but weak positive agreement. Observed agreement was approximately 0.66--0.67, while $P_+$ was only about 0.15--0.16 and $\kappa$ was slightly negative (Table~\ref{tab:iaa-ic-contingency}). Most agreement therefore came from shared non-insertion decisions rather than consistent insertion of implicit intermediate conclusions. Implicit nodes are useful for adjudicated structural representation, but should be treated as a low-reliability layer.

Directed-edge agreement was also weak under strict one-to-one pairing. In the all-pairs setting, one-to-one edit and semantic pairings yielded low positive agreement ($P_+=0.06$ to $0.16$) and low $\kappa$ ($0.02$ to $0.12$). Relaxed same-label pairing improved positive agreement ($P_+=0.39$) and $\kappa$ ($0.36$), suggesting that span granularity explains some structural disagreement. Even so, individual directed edges should not be treated as a high-confidence layer without adjudication (Appendix~\ref{sec:direct_edge_agreement}).

\subsubsection{Directed-path agreement}
Directed-path agreement evaluates whether annotators preserved directed reachability between paired explicit units, even when they chose different local edge decompositions. We report two criteria. The full-transitive criterion counts a target as reachable from a source if any directed path exists between paired units. The stricter implicit-bridge criterion permits traversal through implicit nodes and explicit nodes within the same aligned comparison group, but stops at a different paired explicit unit. This captures source-target recovery through implicit mediation without allowing arbitrary explicit nodes to serve as intermediaries.

\paragraph{Results.}
Directed-path agreement exceeded directed-edge agreement. Full-transitive agreement was strongest under relaxed same-label pairing, with $P_+$ approximately 0.64 and $\kappa$ approximately 0.59 (Table~\ref{tab:iaa_path_summary_revised}). Implicit-bridge agreement was lower, with the strongest same-label relaxed condition yielding $P_+=0.42$ and $\kappa=0.37$. These results show that annotators agreed more often on broad support reachability than on precise local edge placement. The path layer is therefore useful as a structural diagnostic, but individual edges and implicit insertions should still be interpreted cautiously (Appendix~\ref{sec:path_agreement}).

\section{Experiments and Results}

\begin{table*}[t]
\centering
\setlength{\tabcolsep}{3pt}
\renewcommand{\arraystretch}{0.98}

{\small
\begin{tabular*}{\textwidth}{@{\extracolsep{\fill}} %
  l
  S[table-format=1.2] S[table-format=1.2] S[table-format=1.2]
  S[table-format=1.2] S[table-format=1.2] S[table-format=1.2]
  S[table-format=1.2] S[table-format=1.2] S[table-format=1.2]
  S[table-format=1.2] S[table-format=1.2] @{}}
\toprule
& \multicolumn{6}{c}{5 classes}
& \multicolumn{5}{c}{4 classes} \\
\cmidrule(lr){2-7}
\cmidrule(lr){8-12}
Model
  & \multicolumn{1}{c}{\makecell{Macro Avg}}
  & \multicolumn{1}{c}{Analysis}
  & \multicolumn{1}{c}{\makecell{BF}}
  & \multicolumn{1}{c}{Conclusion}
  & \multicolumn{1}{c}{\makecell{PH}}
  & \multicolumn{1}{c}{Rule}
  & \multicolumn{1}{c}{\makecell{Macro Avg}}
  & \multicolumn{1}{c}{Analysis}
  & \multicolumn{1}{c}{\makecell{BF}}
  & \multicolumn{1}{c}{\makecell{PH}}
  & \multicolumn{1}{c}{Rule} \\
\midrule

TF\mbox{-}IDF
  & 0.69 & 0.75 & 0.76 & 0.42 & \textbf{0.82} & 0.69
  & 0.78 & 0.81 & 0.80 & 0.79 & 0.70 \\

SBERT
  & 0.65 & 0.74 & 0.72 & 0.35 & 0.69 & 0.73
  & 0.74 & 0.81 & 0.70 & 0.72 & 0.74 \\

Legal\mbox{-}BERT
  & \textbf{0.71} & \textbf{0.77} & \textbf{0.81}
  & \textbf{0.45} & 0.79 & \textbf{0.75}
  & \textbf{0.80} & \textbf{0.83} & \textbf{0.81}
  & \textbf{0.83} & \textbf{0.75} \\

Modern\mbox{-}BERT
  & 0.65 & 0.73 & 0.78 & 0.39 & 0.65 & 0.70
  & 0.71 & 0.79 & 0.76 & 0.60 & 0.69 \\

\midrule

\makecell[l]{GPT\mbox{-}5\mbox{-}mini$^\dagger$\\
             {\scriptsize with context}}
  & 0.76 & 0.74 & 0.71 & 0.69 & 0.85 & 0.80
  & 0.79 & 0.81 & 0.69 & 0.85 & 0.81 \\

\makecell[l]{GPT\mbox{-}5.4$^\ddagger$\\
             {\scriptsize with context}}
  & 0.79 & 0.79 & 0.81 & 0.67 & 0.89 & 0.80
  & 0.84 & 0.84 & 0.81 & 0.90 & 0.80 \\

\makecell[l]{GPT\mbox{-}5\mbox{-}mini$^\dagger$\\
             {\scriptsize no context}}
  & 0.70 & 0.67 & 0.58 & 0.59 & 0.83 & 0.82
  & 0.74 & 0.75 & 0.55 & 0.85 & 0.80 \\

\makecell[l]{GPT\mbox{-}5.4$^\ddagger$\\
             {\scriptsize no context}}
  & 0.75 & 0.74 & 0.70 & 0.59 & 0.88 & 0.81
  & 0.80 & 0.83 & 0.68 & 0.87 & 0.82 \\

\midrule

Random
  & 0.17 & 0.29 & 0.10 & 0.09 & 0.10 & 0.27
  & 0.21 & 0.34 & 0.12 & 0.13 & 0.26 \\

Majority
  & 0.13 & 0.66 & 0.00 & 0.00 & 0.00 & 0.00
  & 0.18 & 0.71 & 0.00 & 0.00 & 0.00 \\

\bottomrule
\end{tabular*}
}

\caption{Classification experiments with five and four classes (F1-score). Macro Avg denotes macro-averaged F1. Non-GPT classifiers use five-fold case-disjoint \texttt{StratifiedGroupKFold} evaluation. Bold values mark the strongest non-GPT score in each column. $^\dagger$GPT-5-mini uses high reasoning effort. $^\ddagger$GPT-5.4 uses medium reasoning effort. The GPT rows are zero-shot evaluations and all use label definitions. The \textit{with context} condition additionally provides the full case text without annotation labels, whereas the \textit{no context} condition provides only the target passage. BF=\textit{Background Facts}; PH=\textit{Procedural History}.}
\label{tab:classification_experiment}
\end{table*}
\subsection{Classification Experiments}

Explicit passages are classified into functional roles using two settings: a five-label scheme and a four-class scheme that maps \textit{Conclusion} to \textit{Analysis}. Each instance has one adjudicator-selected explicit span as its target and one functional-role label as its output. We report both settings because conclusion passages represent the final step of the court's analysis and can be difficult for a classifier without case context to distinguish from \textit{Analysis}. Implicit intermediate conclusions are excluded, yielding 718 explicit passages from 42 cases.

Non-GPT classifiers employ five-fold case-disjoint \texttt{StratifiedGroupKFold} cross-validation \citep{stone1974crossvalidatory,kohavi1995study,roberts2017crossvalidation}. Passages are grouped by case identifier to ensure that no case appears in both training and test data within a fold. The five-class and four-class folds were generated independently. For each setting, predictions from the five held-out folds are concatenated prior to computing Macro-F1 and per-class F1 scores.

TF-IDF and the three dense representations are classified using a linear support-vector classifier \citep{boser1992training,cortes1995support}. To account for class imbalance, the classifier assigns higher weights to samples from less frequent classes. A fixed random seed ensures reproducibility, and up to 5,000 optimization iterations are allowed. TF-IDF \citep{salton1988term} utilizes word unigrams and bigrams, including features present in at least two training passages and excluding those found in more than 90\% of passages. The dense representations are SBERT \citep{reimers-gurevych-2019-sentence}, LegalBERT \citep{chalkidis-etal-2020-legal}, and ModernBERT \citep{warner-etal-2025-smarter}. Each dense embedding is scaled to unit Euclidean length before classification.

We also report zero-shot GPT-5-mini with high reasoning effort and GPT-5.4 with medium reasoning effort. Both receive label definitions. The with-context condition also receives the full case text without annotation labels, while the no-context condition receives only the target passage. The exact prompt appears in Appendix~\ref{sec:classification_prompt}. These runs are context ablations and points of comparison, not strict baselines.

Table~\ref{tab:classification_experiment} presents Macro-F1 and per-class F1 scores. Merging Conclusion into \textit{Analysis} increases Macro-F1 for all models. Among embedding-based models, LegalBERT achieves the highest performance, with Macro-F1 scores of 0.71 for five classes and 0.80 for four classes. With case context, GPT-5.4 achieves the highest overall Macro-F1, with scores of 0.79 and 0.84, respectively; without context, its scores decrease to 0.75 and 0.80. GPT-5-mini's performance declines from 0.76 and 0.79 with context to 0.70 and 0.74 without context. The largest context effects are observed for \textit{Background Facts}, while \textit{Rule} exhibits minimal change. TF-IDF remains competitive at 0.69 and 0.78 Macro-F1 and achieves 0.82 F1 on five-class \textit{Procedural History}, reflecting stable lexical cues for that role.

\subsection{Retrieval Experiments}

\paragraph{Task.}
We evaluate masked-slot argument retrieval: given an incomplete argument, the retriever must rank candidate sentences so that the missing supporting sentence or sentences appear near the top. The task is multi-positive because each query has one masked slot that may correspond to multiple gold sentences. The dataset comprises 490 queries and 5,286 passages from 42 cases; each query has between 1 and 10 gold sentences, with an average of 2.41.

We employ five case-disjoint outer folds. In each rotation, one fold serves as the test set, the next as the validation set, and the remaining three as the training set. Each case is evaluated once as held-out test data (Appendix~\ref{sec:case_disjoint_fold}). Each query masks a single slot in the argument tree. For internal slots, the builder constructs a maximal consecutive same-label block over the ordered direct premises of a conclusion. This block may include both explicit and implicit nodes. All eligible explicit nodes within the block are hidden. Consecutive refers to the order in the direct-premise list, not necessarily to adjacent text in the case. Any sentence with nonzero character overlap with a hidden explicit node is included in the gold set. The same masking and projection mechanism is applied when the hidden node is a terminal \textit{Conclusion}. Implicit nodes may appear in the visible query context but are not considered retrievable candidates. 

%Appendix~\ref{sec:multi_positive_retrieval_example} provides examples of both sources of multi-positive targets.

\paragraph{Candidate pools.}
All well-formed sentences in the relevant cases are retrievable candidates, not only those overlapping annotated spans. Three pools were evaluated: \textit{Same-case filtered}, which retains the current query's gold sentences but removes same-case passages that are gold only for other queries; \textit{Same-case full}, which ranks all same-case passages; and \textit{Fold-global}, which ranks all passages in the held-out fold. Their mean candidate-pool sizes are 109.4, 134.8, and 1,057.2, respectively. 

%Additionally, a context-excluded Fold-global robustness pool is evaluated, which removes visible non-gold passages without rescoring and always retains gold passages.

\paragraph{Query structure.}
Three query views are employed, although not all are fully crossed with every retriever. The \textit{structured} view preserves tree markup, including the argument root, tree context, focused reasoning step, functional-role tags, and one \texttt{[MASK]} token (Appendix~\ref{sec:structured_query}). The \textit{flat-masked} view linearizes the same content while preserving \texttt{[MASK]} (Appendix~\ref{sec:flat_query_structure}), whereas the flat-plain view replaces the mask with the textual placeholder \texttt{missing span}. BM25 and E5-base-v2 use \textit{flat-plain}. The fixed ModernBERT-base baseline and the fine-tuned flat retriever use \textit{flat-masked}, while the fine-tuned structured retriever uses the \textit{structured} view. BM25 uses Pyserini/Lucene with (k\_1=0.9) and (b=0.4). E5 uses \texttt{intfloat/e5-base-v2} with the standard \texttt{query:} and \texttt{passage:} prefixes and a deterministic procedure that prioritizes the focused reasoning step when fitting queries within its 512-position limit.

\paragraph{Training.}
The fine-tuned retrievers are ModernBERT dual encoders initialized from \texttt{answerdotai/ModernBERT-base}. The query encoder uses the \texttt{[MASK]} hiddenstate, while the passage encoder uses normalized mean pooling over non-padding tokens, excluding the first special token. Both encoders are trained with a multi-positive contrastive objective, using log-sum-exp over positives in the numerator and over all candidates in the denominator.

Flat and structured retrievers share the same setup and differ only in query view. Training is conducted for 20 epochs with a learning rate of $10^{-5}$, temperature 0.07, maximum query length 4096, and maximum passage length 500. Each example uses up to four positives and one of two negative samplers. \textit{Case-focused} training samples 40 unique negatives from the query's case and 20 from other training cases; \textit{pool-uniform} training samples 60 unique negatives passage-uniformly from the training folds. Positive selection remains constant across query views and samplers. Both query views are trained with both samplers and seeds 17, 29, and 43 in every fold. Checkpoints are selected to maximize validation case-macro Recall@20, then untruncated MRR, with the earlier epoch breaking a tie (Appendix~\ref{sec:retrieval_training_setup}).

\begin{figure}[!ht]
\centering

\pgfplotsset{
  flat/.style={ftflat},
  structured/.style={ftstruct},
  localunique/.style={
    solid,
    mark=*
  },
  globaluniform/.style={
    densely dashed,
    mark=square*
  },
  random/.style={
    black!70,
    densely dashed,
    mark=none,
    line width=0.8pt
  }
}

\begin{tikzpicture}

\begin{groupplot}[
  group style={
    group size=1 by 3,
    vertical sep=1.5cm,
    xlabels at=edge bottom,
    ylabels at=edge left
  },
  width=0.95\columnwidth,
  height=0.7\columnwidth,
  xmin=1,
  xmax=20,
  ymin=0,
  ymax=85,
  xtick={2,4,6,8,10,12,14,16,18,20},
  ytick={0,20,40,60,80},
  xlabel={$k$},
  ylabel={Case-macro Hit@$k$},
  grid=both,
  grid style={black!18, densely dotted},
  tick label style={font=\footnotesize},
  label style={font=\footnotesize},
  title style={font=\footnotesize},
  tick style={black!60},
  tick align=outside,
  scaled y ticks=false,
  yticklabel={\pgfmathprintnumber{\tick}\%}
]

\nextgroupplot[
  title={Same-case filtered},
  legend to name=retrievallegend,
  legend columns=2,
  legend cell align={left},
  legend style={
    draw=none,
    fill=none,
    font=\small,
    column sep=0.6em,
    /tikz/every even column/.append style={column sep=0.8em}
  }
]

\addplot+[bm25] coordinates {
  (2,22.919500)
  (4,38.282304)
  (6,46.984414)
  (8,52.304745)
  (10,58.961793)
  (12,64.304291)
  (14,67.767548)
  (16,69.506770)
  (18,71.617520)
  (20,73.920238)
};
\addlegendentry{BM25}

\addplot+[e5] coordinates {
  (2,13.817042)
  (4,27.597907)
  (6,37.441806)
  (8,46.045559)
  (10,52.395610)
  (12,56.878673)
  (14,61.362595)
  (16,63.915110)
  (18,66.414210)
  (20,68.736746)
};
\addlegendentry{E5}

\addplot+[mbbase] coordinates {
  (2,6.435710)
  (4,13.230056)
  (6,18.421195)
  (8,27.425954)
  (10,32.258002)
  (12,35.334336)
  (14,38.363579)
  (16,42.077082)
  (18,45.690186)
  (20,49.188498)
};
\addlegendentry{ModernBERT-base}

\addplot+[random] coordinates {
  (2,6.190937)
  (4,11.902654)
  (6,17.184799)
  (8,22.081282)
  (10,26.630947)
  (12,30.868170)
  (14,34.823382)
  (16,38.523533)
  (18,41.992511)
  (20,45.251496)
};
\addlegendentry{Random}

\addplot+[flat,localunique] coordinates {
  (2,22.633741)
  (4,38.611779)
  (6,48.633304)
  (8,57.347864)
  (10,63.152102)
  (12,68.645330)
  (14,73.031380)
  (16,75.692246)
  (18,78.121191)
  (20,80.449985)
};
\addlegendentry{Flat/case-focused}

\addplot+[flat,globaluniform] coordinates {
  (2,22.275883)
  (4,39.059944)
  (6,47.720310)
  (8,56.268184)
  (10,63.822811)
  (12,68.154705)
  (14,72.484416)
  (16,75.438178)
  (18,77.396083)
  (20,79.941405)
};
\addlegendentry{Flat/pool-uniform}

\addplot+[structured,localunique] coordinates {
  (2,24.643013)
  (4,36.759477)
  (6,44.298398)
  (8,49.853482)
  (10,56.991661)
  (12,62.924328)
  (14,68.398562)
  (16,72.341258)
  (18,75.674060)
  (20,78.431671)
};
\addlegendentry{Struct./case-focused}

\addplot+[structured,globaluniform] coordinates {
  (2,25.079886)
  (4,38.307438)
  (6,45.033352)
  (8,50.994429)
  (10,58.349255)
  (12,62.959696)
  (14,67.951092)
  (16,71.655403)
  (18,75.310838)
  (20,78.501208)
};
\addlegendentry{Struct./pool-uniform}

\nextgroupplot[
  title={Same-case full}
]

\addplot+[bm25] coordinates {
  (2,2.948128)
  (4,7.756726)
  (6,14.198535)
  (8,22.619866)
  (10,28.615744)
  (12,35.260761)
  (14,41.146519)
  (16,48.696047)
  (18,52.159732)
  (20,55.206803)
};

\addplot+[e5] coordinates {
  (2,3.260258)
  (4,11.622097)
  (6,18.361433)
  (8,24.687425)
  (10,32.818746)
  (12,39.482639)
  (14,44.647434)
  (16,48.682751)
  (18,53.106427)
  (20,56.410669)
};

\addplot+[mbbase] coordinates {
  (2,5.210199)
  (4,8.048197)
  (6,13.071027)
  (8,18.858236)
  (10,25.400496)
  (12,29.398454)
  (14,34.277250)
  (16,36.168629)
  (18,37.759962)
  (20,40.145617)
};

\addplot+[random] coordinates {
  (2,4.849214)
  (4,9.405493)
  (6,13.692907)
  (8,17.733274)
  (10,21.546373)
  (12,25.150149)
  (14,28.560882)
  (16,31.793354)
  (18,34.860993)
  (20,37.776003)
};

\addplot+[flat,localunique] coordinates {
  (2,12.652105)
  (4,23.378566)
  (6,32.111813)
  (8,40.444634)
  (10,46.417132)
  (12,51.897052)
  (14,56.009951)
  (16,59.873143)
  (18,64.029683)
  (20,67.978451)
};

\addplot+[flat,globaluniform] coordinates {
  (2,11.676942)
  (4,23.155946)
  (6,33.080280)
  (8,40.003369)
  (10,46.508655)
  (12,51.111656)
  (14,56.211644)
  (16,60.416982)
  (18,63.951231)
  (20,66.982640)
};

\addplot+[structured,localunique] coordinates {
  (2,13.760669)
  (4,22.745962)
  (6,32.522499)
  (8,38.354991)
  (10,44.191593)
  (12,48.660647)
  (14,51.720248)
  (16,55.079448)
  (18,58.837430)
  (20,62.699409)
};

\addplot+[structured,globaluniform] coordinates {
  (2,12.668082)
  (4,23.092920)
  (6,32.973634)
  (8,39.499775)
  (10,45.065447)
  (12,48.291868)
  (14,52.011631)
  (16,55.381395)
  (18,58.984725)
  (20,63.003261)
};

\nextgroupplot[
  title={Fold-global}
]

\addplot+[bm25] coordinates {
  (2,2.778060)
  (4,7.015985)
  (6,12.454789)
  (8,16.272816)
  (10,22.139123)
  (12,26.697352)
  (14,32.467172)
  (16,37.078253)
  (18,39.963106)
  (20,42.485491)
};

\addplot+[e5] coordinates {
  (2,2.605496)
  (4,8.052845)
  (6,14.928770)
  (8,18.927990)
  (10,23.375675)
  (12,26.673475)
  (14,28.552889)
  (16,30.885010)
  (18,33.869749)
  (20,35.650964)
};

\addplot+[mbbase] coordinates {
  (2,0.340136)
  (4,0.879792)
  (6,1.482178)
  (8,2.201043)
  (10,2.498662)
  (12,2.681812)
  (14,2.800860)
  (16,3.140996)
  (18,4.018590)
  (20,4.677931)
};

\addplot+[random] coordinates {
  (2,0.492416)
  (4,0.982189)
  (6,1.469335)
  (8,1.953872)
  (10,2.435818)
  (12,2.915190)
  (14,3.392004)
  (16,3.866278)
  (18,4.338027)
  (20,4.807270)
};

\addplot+[flat,localunique] coordinates {
  (2,1.742907)
  (4,3.309666)
  (6,4.747582)
  (8,6.742671)
  (10,8.977222)
  (12,10.267542)
  (14,12.014933)
  (16,13.191526)
  (18,14.438606)
  (20,15.672040)
};

\addplot+[flat,globaluniform] coordinates {
  (2,1.440709)
  (4,3.170533)
  (6,4.585242)
  (8,7.013829)
  (10,9.016966)
  (12,11.108213)
  (14,12.899112)
  (16,13.894862)
  (18,14.772178)
  (20,15.584534)
};

\addplot+[structured,localunique] coordinates {
  (2,2.257386)
  (4,3.330618)
  (6,4.606324)
  (8,6.304541)
  (10,7.556961)
  (12,8.779944)
  (14,9.493213)
  (16,10.734682)
  (18,11.644342)
  (20,12.996220)
};

\addplot+[structured,globaluniform] coordinates {
  (2,2.087729)
  (4,3.123066)
  (6,4.573913)
  (8,6.289526)
  (10,7.233241)
  (12,8.411096)
  (14,9.402979)
  (16,10.458437)
  (18,12.033069)
  (20,13.222023)
};

\end{groupplot}

\end{tikzpicture}

\vspace{0.2em}

\makebox[\columnwidth][c]{%
  \pgfplotslegendfromname{retrievallegend}%
}

\caption{Case-macro Hit@$k$ (\%) across three candidate-pool regimes using five-fold case-disjoint evaluation. For each fine-tuned model, the three seed-specific case means are averaged, then averaged equally across all cases. Same-case filtered excludes gold passages for other queries; Same-case full includes all passages from the query’s case; Fold-global ranks all passages in the held-out fold. \textit{Case-focused} refers to 40 same-case negatives and 20 from other cases, while the \textit{pool-uniform} curve corresponds to 60 negatives uniformly sampled from the complete training fold.
}
\label{fig:retrieval_curves}

\end{figure}
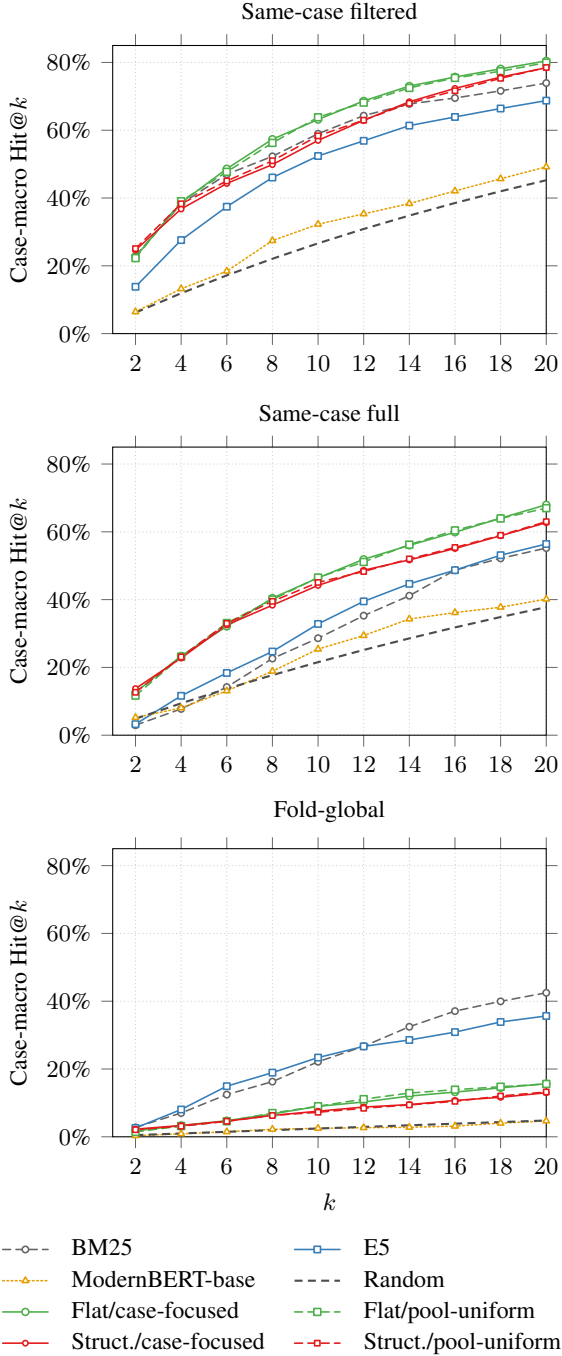

\paragraph{Evaluation.}
We use case-macro Fold-global Hit@20 as the primary evaluation metric. Query-level metrics are first averaged within each held-out case. For fine-tuned systems, the three seed-specific case means are then averaged before weighting all 42 cases equally. Hit@\(K\) measures whether at least one gold sentence appears in the top \(K\); Recall@\(K\) measures the fraction of each query's gold set recovered; Complete recovery@\(K\) requires all gold sentences to appear in the top \(K\); and untruncated MRR uses the rank of the first gold sentence in the complete candidate-pool ranking. Table~\ref{tab:retrieval_cv_metrics_at_20} reports the metrics at \(K=20\), and Figure~\ref{fig:retrieval_curves} plots Hit@\(K\) at intervals of two. The random baseline is the case-macro analytic probability that a uniformly random top-\(K\) sample contains at least one gold sentence. To determine the sensitivity of retriever comparisons to the composition of the 42-case test collection, paired case-resampling intervals are calculated using 10,000 bootstrap samples of the fixed per-case results.

\paragraph{Results.} 
Results vary by candidate pool. In the \textit{Same-case filtered} regime, the fine-tuned flat retriever with case-focused sampling performs best on the main multi-positive metrics, reaching 80.45\% Hit@20, 57.17\% Recall@20, and 36.04\% Complete recovery@20. Both fine-tuned representations improve substantially over ModernBERT-base in this regime.

In the \textit{Same-case full} regime, the fine-tuned flat retriever with case-focused sampling again performs best on the main multi-positive metrics, with 67.98\% Hit@20, 43.98\% Recall@20, and 26.07\% Complete recovery@20. The fine-tuned structured retriever with case-focused sampling has the highest MRR at 18.87\%, indicating a slightly earlier first hit, but recovers less of the complete gold set.

In the \textit{Fold-global} regime, BM25 leads all four metrics, with 42.49\% Hit@20, 23.99\% Recall@20, 12.82\% Complete recovery@20, and 7.84\% MRR. E5-base-v2 follows with 35.65\% Hit@20. The fine-tuned ModernBERT retrievers are weaker: the flat models reach 15.67\% with case-focused sampling and 15.58\% with pool-uniform sampling. Overall, fine-tuning helps in the within-case settings but does not show robust cross-case generalization, and tree structure does not provide a consistent advantage over the flat representation (Appendix~\ref{sec:retrieval_expanded_results}).

All 95\% paired case-bootstrap intervals for the prespecified Fold-global Hit@20 comparisons included zero, providing no reliable evidence of an advantage from structured queries or pool-uniform negative sampling (Appendix~\ref{sec:retrieval_uncertainty}).

\section{Discussion}

Explicit functional node labels represent the most reliable component of the resource. Span-level and sentence-level agreement are nearly identical, while implicit intermediate conclusions and direct support edges exhibit less stability. Path agreement exceeds direct-edge agreement, indicating that annotators frequently maintain similar broad reachability relations even when decomposing local support differently. Annotator A1 inserted 11 implicit nodes, whereas A2 inserted 30, suggesting that the annotators employed different thresholds for determining when an implicit step was strictly necessary (Appendix~\ref{sec:iic_per_annotator}). Consequently, the functional node-label layer should be regarded as the primary reliable annotation layer. The tree layer is best considered an exploratory representation, with directed paths offering a more stable structural benchmark than individual edges.

The classification results show that functional labels are learnable under case-disjoint evaluation, although the five-class functional-role scheme remains challenging. The main confusion occurs between \textit{Rule} and \textit{Analysis}. The four-class setting yields higher Macro-F1 scores across all models. The GPT context experiments demonstrate that context is particularly important for Background Facts, while performance on Rule is largely unaffected by the removal of case context.

The retrieval experiments show a within-case benefit from supervised fine-tuning, but no consistent advantage for structured queries over flat queries. The flat retriever achieves the highest performance on the main set-recovery metrics in both same-case settings, while the structured retriever attains a slightly higher MRR. In the \textit{Fold-global} setting, BM25 performs best, and both fine-tuned representations generalize weakly across cases. The paired case-bootstrap intervals do not provide reliable evidence that structured queries or pool-uniform negative sampling improve Fold-global Hit@20.

\section{Conclusion}

We introduced an expert-annotated dataset of 42 U.S. federal tax opinions on corporate reorganizations under I.R.C.\ \S368. The corpus labels spans by function: \textit{Rule}, \textit{Analysis}, \textit{Conclusion}, \textit{Background Facts}, and \textit{Procedural History}, and provides span-based, sentence-based, flat, and tree-structured representations of chained support relations.

Under case-disjoint evaluation, functional labels are learned reliably, while structural annotations are less reproducible, suggesting that argumentative reachability is more stable than precise local decomposition. The functional layer should therefore be treated as the corpus's primary representation, with the tree layer serving as an exploratory structural benchmark.

Retrieval fine-tuning improves within-case ranking over ModernBERT-base, but BM25 remains stronger under fold-global evaluation. Together, these results position the dataset as a benchmark for both legal passage classification and the study of how reliably structured legal arguments can be annotated and generalized across cases.

\section*{Limitations} % FYI this does not count toward the total page limit

The corpus comprises 42 cases from a specialized area of English-language U.S. federal tax law. Due to the high cost of expert annotation, both the corpus size and the quantity of supervised data available for model training are limited. The results may not generalize to other tax doctrines, legal domains, jurisdictions, or non-legal texts.

The opinions span from 1935 to 1987. As the corpus lacks post-1990 opinions, it is not temporally comprehensive and does not reflect recent case language or developments in corporate reorganization practice. Future work should incorporate modern opinions and directly assess temporal transfer.

The reliability of the annotation layers varies. For each case, the adjudicator selected a single complete annotation file without editing or merging individual components. As a result, the released trees represent coherent, expert-selected interpretations, but should not be regarded as the sole valid local decompositions.

The retrieval task employs the selected trees to define hidden slots and gold passages, measuring recoverability under the released representation. Fold-global results show weak cross-case transfer for fine-tuned retrievers, and the tested tree markup does not solve that problem. Close differences between query representations and negative samplers are considered descriptive, as their paired case-bootstrap intervals include zero.

\section*{Acknowledgments}

ChatGPT and Grammarly were utilized to assist with language refinement and LaTeX formatting. All generated content was reviewed and revised by the authors, who assume full responsibility for the final manuscript. 

%\section*{Acknowledgments}

\bibliography{custom}

\clearpage

\appendix

\clearpage

\twocolumn[{
\begin{minipage}{\textwidth}

\section{Annotation Structure Example} 
\label{sec:annotation_structure}
\captionsetup{type=figure}

\vspace{0.08in}

\centering

\begin{tikzpicture}[
  trim left=0pt,
  trim right=\linewidth,
  font=\scriptsize,
  box/.style={
    draw,
    rectangle,
    rounded corners=1.4pt,
    align=left,
    inner sep=2.2pt,
    outer sep=0pt
  },
  concl/.style={
    box,
    fill=red!20,
    text width=.68\linewidth
  },
  analysisMain/.style={
    box,
    fill=blue!10,
    text width=.42\linewidth
  },
  analysisWide/.style={
    box,
    fill=blue!10,
    text width=.25\linewidth
  },
  analysis/.style={
    box,
    fill=blue!10,
    text width=.26\linewidth
  },
  analysisSmall/.style={
    box,
    fill=blue!10,
    text width=.21\linewidth
  },
  ruleWide/.style={
    box,
    fill=green!20,
    text width=.34\linewidth
  },
  rule/.style={
    box,
    fill=green!20,
    text width=.26\linewidth
  },
  ruleSmall/.style={
    box,
    fill=green!20,
    text width=.21\linewidth
  },
  implicit/.style={
    box,
    fill=green!20,
    text width=.28\linewidth
  },
  context/.style={
    box,
    text width=.40\linewidth
  },
  ph/.style={
    context,
    fill=yellow!20
  },
  bf/.style={
    context,
    fill=purple!18
  },
  link/.style={
    -{Latex[length=1.5mm]},
    semithick
  }
]

% ============================================================
\path
  (0,0) coordinate (pageLeft)
  -- (\linewidth,0) coordinate (pageRight);

% ============================================================

% Root
\coordinate (xC) at (.50\linewidth,0);

% Level 1
\coordinate (xA1)   at (.30\linewidth,0);
\coordinate (xRtop) at (.78\linewidth,0);

% Level 2
\coordinate (xR368) at (.21\linewidth,0);
\coordinate (xA368) at (.70\linewidth,0);

% Level 3
\coordinate (xAWY)   at (.26\linewidth,0);
\coordinate (xRcont) at (.54\linewidth,0);
\coordinate (xAXOk)  at (.82\linewidth,0);

% Level 4
\coordinate (xR677) at (.14\linewidth,0);
\coordinate (xAZD)  at (.38\linewidth,0);

% Contextual annotations
\coordinate (xPH) at (.26\linewidth,0);
\coordinate (xBF) at (.74\linewidth,0);

% ============================================================
\node[concl, anchor=north] (C)
  at (xC) {%
\textbf{Conclusion:}
The transactions that occurred in the instant case, which in substance
were really a continuation of the insurance business rather than its
cessation, were properly characterized as a reorganization.
\par\medskip
\textbf{Affirmed.}
};

% ============================================================

\coordinate (L1)
  at ([yshift=-5mm]C.south);

\node[analysisMain, anchor=north] (A1)
  at (xA1 |- L1) {%
\textbf{Analysis:}
Under these circumstances, it appears that Ringwalt should be treated
as the owner of the Clifford Trust, pursuant to section 677(a)(2),[10]
and correspondingly we hold that the common control requirement for a
reorganization, defined by section 368, was satisfied.
};

\node[implicit, anchor=north] (Rtop)
  at (xRtop |- L1) {%
\textbf{Rule / Intermediate Implicit Conclusion}
};

% ============================================================

\coordinate (L2)
  at ([yshift=-5mm]current bounding box.south);

\node[ruleWide, anchor=north] (R368)
  at (xR368 |- L2) {%
\textbf{Rule:}
I.R.C.\ \S 368(a)(1)(D), which defines a reorganization as:
a transfer by a corporation of all or a part of its assets to another
corporation if immediately after the transfer the transferor, or one
or more of its shareholders ``\ldots'' assets are transferred are
distributed in a transaction which qualifies under section 354, 355,
or 356.
};

\node[analysisWide, anchor=north] (A368)
  at (xA368 |- L2) {%
\textbf{Analysis:}
The series of transactions that took place in the instant case appears
governed by I.R.C.\ \S 368(a)(1)(D).
};

% ============================================================

\coordinate (L3)
  at ([yshift=-6mm]current bounding box.south);

\node[analysisWide, anchor=north] (AWY)
  at (xAWY |- L3) {%
\textbf{Analysis:}
The only beneficial right that Ringwalt relinquished under the trust
agreement was the right to receive trust receipts allocable to income.
Ringwalt had numerous powers of administration ``\ldots'' extensive
power to allocate trust receipts between principal and income.
};

\node[rule, anchor=north] (Rcont)
  at (xRcont |- L3) {%
\textbf{Rule:}
Assessing continuity of interest ultimately depends upon proof of
beneficial ownership without regard to the existence or absence of
legal title. See \textit{Bondy v.\ Commissioner}, 269 F.2d 463,
466--67 (4th Cir.\ 1959).
};

\node[analysis, anchor=north] (AXOk)
  at (xAXOk |- L3) {%
\textbf{Analysis:}
In determining that the continuity of interest requirement had been
established in the instant case, the district court specifically found
that Ringwalt was treated appropriately as the owner of 84\% of the
R \& L, Inc.\ stock because ``\ldots''.
};

% ============================================================

\coordinate (L4)
  at ([yshift=-5mm]current bounding box.south);

\node[ruleSmall, anchor=north] (R677)
  at (xR677 |- L4) {%
\textbf{Rule:}
I.R.C.\ \S 677(a)(2) provides that the grantor ``\ldots''.
};

\node[analysisSmall, anchor=north] (AZD)
  at (xAZD |- L4) {%
\textbf{Analysis:}
In accordance with Ringwalt's powers as trustee, ``\ldots''
};

% ============================================================

% Level 1 -> root
\draw[link]
  (A1.north) -- ([xshift=-10mm]C.south);

\draw[link]
  (Rtop.north) -- ([xshift=10mm]C.south);

% Level 2 -> level 1
\draw[link]
  (R368.north) -- ([xshift=-10mm]A1.south);

\draw[link]
  (A368.north) -- ([xshift=10mm]A1.south);

% Level 3 -> A368
\draw[link]
  (AWY.north) -- ([xshift=-12mm]A368.south);

\draw[link]
  (Rcont.north) -- (A368.south);

\draw[link]
  (AXOk.north) -- ([xshift=12mm]A368.south);

% Level 4 -> AWY
\draw[link]
  (R677.north) -- ([xshift=-8mm]AWY.south);

\draw[link]
  (AZD.north) -- ([xshift=8mm]AWY.south);

% ============================================================

\coordinate (ContextLevel)
  at ([yshift=-3mm]current bounding box.south);

\node[ph, anchor=north] (PH1)
  at (xPH |- ContextLevel) {%
\textbf{Procedural History:}
Jack D.\ Ringwalt and other taxpayers[1] appeal from the district
court's[2] judgment disallowing their claims for income tax refunds
for the year 1967. ``\ldots''.
};

\node[bf, anchor=north] (BF)
  at (xBF |- ContextLevel) {%
\textbf{Background Facts:}
The basic facts are described in a stipulation adopted by the district
court. ``\ldots'' Ringwalt retained a reversionary interest in the
trust corpus and held various powers.
};

\end{tikzpicture}

\caption{
Condensed view of syllogistic argument tree annotation of the case \textit{Ringwalt v.\ U.S.}, 549 F.2d 89 (8th Cir.\ 1977). \textit{Background Facts} and \textit{Procedural History} are included in the annotation but are not considered part of the argument structure, as they are defined as contextual spans rather than as support for the argument's claims. }
\label{fig:ringwalt_tree}
\vspace{1em}

\section{Annotation Workflow}
\label{sec:annotation_workflow}
\captionsetup{type=figure}
\vspace{0em}

    \centering
\begin{minipage}{1\textwidth}
    \includegraphics[width=1\linewidth]{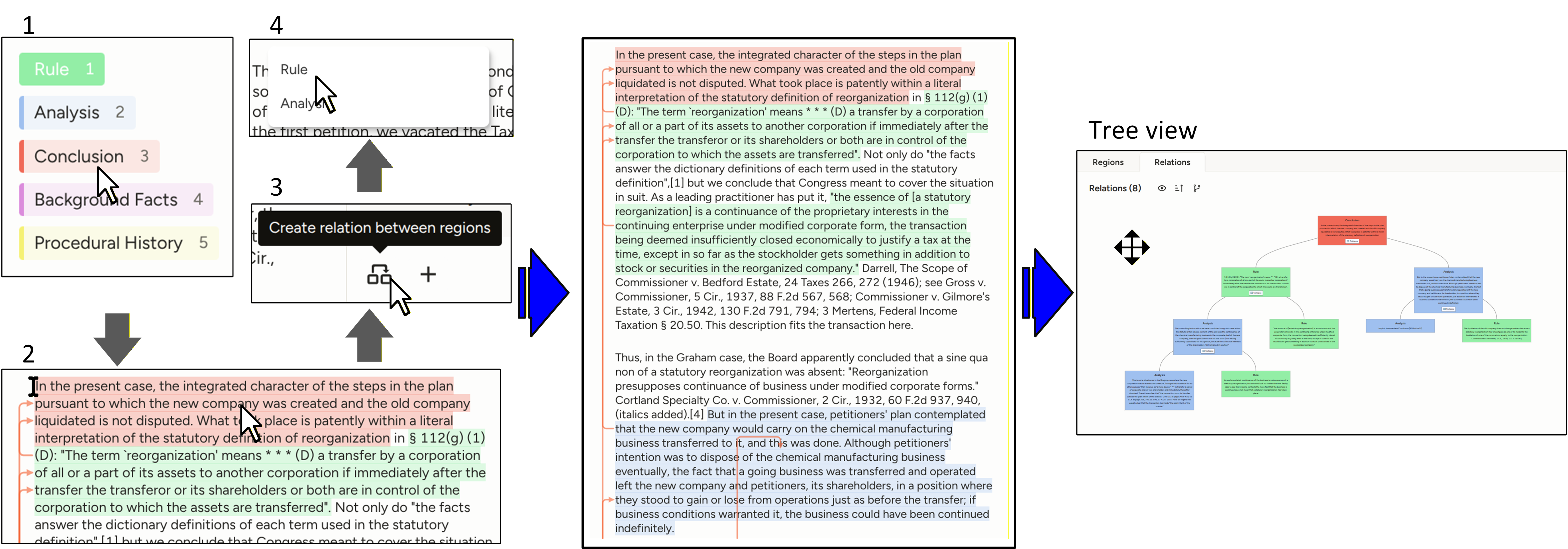}
\end{minipage}
    \caption{Annotation workflow, visualization of the annotation interface, structure tree visualization, and an example of an annotated case data format. The workflow: (1) selecting the label; (2) highlighting the text, which automatically applies the label's color and saves the annotation; (3) creating edges by pressing the 'create relation' button with one span selected, then selecting another annotation; and (4) for Intermediate Implicit Conclusions, right-clicking displays available labels and inserts a new block.}
    \label{fig:annotation_workflow}

\end{minipage}
}]

\clearpage
\section{Dataset Statistics}
\label{sec:dataset_statistics}
\begin{table}[ht]
\centering
\footnotesize
\setlength{\tabcolsep}{2.5pt}
\renewcommand{\arraystretch}{0.96}

\begin{tabular*}{\columnwidth}{@{\extracolsep{\fill}}lr@{}}
\toprule
\multicolumn{2}{@{}l}{\textbf{Corpus and structure summary}} \\
\midrule
Cases & 42 \\
Total words & 150,040 \\
Average words / case & 3,572.38 \\
Median words / case & 2,989.50 \\
Total sentences & 5,286 \\
Average sentences / case & 125.86 \\
Median sentences / case & 105.50 \\
Explicit spans & 718 \\
Average spans / case & 17.10 \\
Median spans / case & 14.50 \\
Nodes / edges & 800 / 644 \\
Argument trees & 43 \\
Implicit insertions & 82 \\
Disconnected spans & 132 \\
Average depth & 2.41 \\
Max depth & 10 \\
Average branching & 2.17 \\
\bottomrule
\end{tabular*}

\vspace{0.45em}

\begin{tabular*}{\columnwidth}{@{\extracolsep{\fill}}lrrrrrr@{}}
\toprule
\multicolumn{1}{@{}l}{} &
\multicolumn{2}{c}{\textbf{Explicit spans}} &
\multicolumn{2}{c}{\textbf{All nodes}} &
\multicolumn{2}{c@{}}{\textbf{Sentences}} \\
\multicolumn{1}{@{}l}{} &
\multicolumn{2}{c}{($n=718$)} &
\multicolumn{2}{c}{($n=800$)} &
\multicolumn{2}{c@{}}{($n=2715$)} \\
\cmidrule(lr){2-3}\cmidrule(lr){4-5}\cmidrule(l){6-7}
Label & Count & \% & Count & \% & Count & \% \\
\midrule
BF & 59 & 8.22 & 59 & 7.38 & 1,420 & 52.30 \\
PH & 56 & 7.80 & 56 & 7.0 & 160 & 5.89 \\
Rule & 206 & 28.69 & 243 & 30.38 & 428 & 15.76 \\
Analysis & 353 & 49.16 & 398 & 49.75 & 620 & 22.84 \\
Conclusion & 44 & 6.13 & 44 & 5.50 & 87 & 3.20 \\
\bottomrule
\end{tabular*}

\caption{Descriptive corpus statistics. Sentence-label percentages are computed over the 2,715 labeled candidate sentences used in sentence-level alignment, not all 5,286 case sentences. BF=\textit{Background Facts}; PH=\textit{Procedural History}.}
\label{tab:dataset_statistics}
\end{table}

\section{Implicit nodes annotation count}
\label{sec:iic_per_annotator}
\begin{table}[!ht]
\centering
\scriptsize
\setlength{\tabcolsep}{3.5pt}
\resizebox{\columnwidth}{!}{%
\begin{tabular}{r l l r r r r r r}
\toprule
\textbf{Ref} & \textbf{File} & \textbf{Annot.} & \textbf{Anal.} & \textbf{Rule} & \textbf{Tot.} & \textbf{AvgA} & \textbf{AvgR} & \textbf{AvgT} \\
\midrule
4  & 35 & A1 & 0 & 0 & 0 & --  & --  & --  \\
4  & 85 & A2 & 7 & 0 & 7 & --  & --  & --  \\
5  & 64 & A1 & 1 & 0 & 1 & --  & --  & --  \\
5  & 63 & A2 & 1 & 1 & 2 & --  & --  & --  \\
6  & 57 & A1 & 1 & 1 & 2 & --  & --  & --  \\
6  & 90 & A2 & 2 & 2 & 4 & --  & --  & --  \\
7  & 72 & A1 & 1 & 0 & 1 & --  & --  & --  \\
7  & 39 & A2 & 1 & 1 & 2 & --  & --  & --  \\
8  & 41 & A1 & 0 & 0 & 0 & --  & --  & --  \\
8  & 84 & A2 & 3 & 1 & 4 & --  & --  & --  \\
9  & 73 & A1 & 1 & 0 & 1 & --  & --  & --  \\
9  & 93 & A2 & 1 & 3 & 4 & --  & --  & --  \\
10 & 36 & A1 & 2 & 1 & 3 & --  & --  & --  \\
10 & 95 & A2 & 1 & 1 & 2 & --  & --  & --  \\
11 & 37 & A1 & 0 & 1 & 1 & --  & --  & --  \\
11 & 52 & A2 & 2 & 0 & 2 & --  & --  & --  \\
12 & 45 & A1 & 2 & 0 & 2 & --  & --  & --  \\
12 & 51 & A2 & 2 & 0 & 2 & --  & --  & --  \\
13 & 69 & A1 & 0 & 0 & 0 & --  & --  & --  \\
13 & 43 & A2 & 1 & 0 & 1 & --  & --  & --  \\
\midrule
\multicolumn{2}{l}{\textbf{Summary}} & A1 & 8  & 3 & 11 & 0.8 & 0.3 & 1.1 \\
\multicolumn{2}{l}{\textbf{Summary}} & A2 & 21 & 9 & 30 & 2.1 & 0.9 & 3.0 \\
\bottomrule
\end{tabular}%
}
\caption{Annotation counts. AvgA and AvgR denote average Analysis and Rule implicit nodes insertion.}
\label{tab:annotation_stats}
\end{table}

\section{Classification task confusion matrices}
\label{sec:confusion_matrices}
\begin{table}[ht]
\centering
\small
\setlength{\tabcolsep}{2.8pt}
\renewcommand{\arraystretch}{0.96}
\begin{tabular*}{\columnwidth}{@{\extracolsep{\fill}}l
  S[table-format=1.2]
  S[table-format=1.2]
  S[table-format=1.2]
  S[table-format=1.2]
  S[table-format=1.2]@{}}
\toprule
Gold & \multicolumn{1}{c}{A} & \multicolumn{1}{c}{BF} & \multicolumn{1}{c}{C} & \multicolumn{1}{c}{PH} & \multicolumn{1}{c}{R} \\
\midrule
\multicolumn{6}{@{}l}{\textbf{Legal-BERT} (Macro-F1 = 0.71)} \\
Analysis   & 0.75 & 0.03 & 0.04 & 0.02 & 0.17 \\
BF         & 0.05 & 0.83 & 0.00 & 0.12 & 0.00 \\
Conclusion & 0.41 & 0.00 & 0.41 & 0.16 & 0.02 \\
PH         & 0.02 & 0.07 & 0.02 & 0.89 & 0.00 \\
Rule       & 0.21 & 0.00 & 0.01 & 0.00 & 0.77 \\
\midrule
\multicolumn{6}{@{}l}{\textbf{Modern-BERT} (Macro-F1 = 0.65)} \\
Analysis   & 0.73 & 0.02 & 0.07 & 0.05 & 0.13 \\
BF         & 0.08 & 0.80 & 0.00 & 0.10 & 0.02 \\
Conclusion & 0.43 & 0.00 & 0.41 & 0.11 & 0.05 \\
PH         & 0.11 & 0.07 & 0.05 & 0.77 & 0.00 \\
Rule       & 0.30 & 0.01 & 0.02 & 0.01 & 0.66 \\
\midrule
\multicolumn{6}{@{}l}{\textbf{GPT-5-mini with context} (Macro-F1 = 0.76)}  \\
Analysis   & 0.65 & 0.08 & 0.03 & 0.00 & 0.24 \\
BF         & 0.08 & 0.86 & 0.00 & 0.03 & 0.02 \\
Conclusion & 0.34 & 0.00 & 0.66 & 0.00 & 0.00 \\
PH         & 0.05 & 0.11 & 0.04 & 0.79 & 0.02 \\
Rule       & 0.05 & 0.00 & 0.00 & 0.00 & 0.95 \\
\bottomrule
\end{tabular*}
\caption{Row-normalized confusion matrices for the five-class case-disjoint experiment. Rows are gold labels and columns are predicted labels. A=Analysis, BF=\textit{Background Facts}, C=Conclusion, PH=\textit{Procedural History}, and R=Rule.}
\label{tab:classification_confusion_5class}
\end{table}

\section{Retrieval Models Training Setup}
\label{sec:retrieval_training_setup}
\paragraph{Dual-encoder scoring.}
The retriever is initialized from ModernBERT-base and uses one shared encoder for queries and passages, with representation-specific pooling. For a query $q$, the query representation is the final-layer hidden state at its single \texttt{[MASK]} token. For a candidate passage $c$, the passage representation is obtained by mean pooling the non-padding hidden states after excluding sequence position zero. Both representations are $L^2$-normalized:

$$ \mathbf z_q = \frac{g_\theta^{q}(q)}{\lVert g_\theta^{q}(q)\rVert_2},
  \qquad
  \mathbf z_c = \frac{g_\theta^{p}(c)}{\lVert g_\theta^{p}(c)\rVert_2}, $$

where $g_\theta^{q}$ and $g_\theta^{p}$ denote the query- and passage-pooling operations applied to the same encoder with parameters $\theta$. Their similarity is

$$s_\theta(q,c)=\mathbf z_q^\top\mathbf z_c.$$

Because both vectors are normalized, this dot product is cosine similarity.

\paragraph{Multi-positive contrastive objective.}
Let $\mathcal B_m$ be the set of real queries in distributed microbatch $m$. For each query $q\in\mathcal B_m$, let $\mathcal A_q$ be its sampled candidate set, containing between one and four explicitly selected positive passages and exactly 60 negative passages. The candidate set shared across the distributed microbatch is

$$\mathcal C_m =
  \bigcup_{q\in\mathcal B_m}\mathcal A_q,$$
  
where the union is taken over passage identifiers across all workers. Consequently, a passage proposed by multiple queries or workers occurs only once in $\mathcal C_m$.

Let $\mathcal P_q$ denote the complete set of gold passage identifiers for query $q$, and let $\tau>0$ be the temperature. The gold passages that contribute to the loss are those present in the current microbatch candidate set, $\mathcal P_q\cap\mathcal C_m$. This intersection is guaranteed to be nonempty because every query explicitly contributes at least one positive candidate. The loss for query $q$ is

$$\ell_{q,m} = -\log \frac{\displaystyle \sum_{c\in\mathcal P_q\cap\mathcal C_m} \exp\left(s_\theta(q,c)/\tau\right)}{\displaystyle \sum_{c\in\mathcal C_m} \exp\left(s_\theta(q,c)/\tau\right)}.$$

\paragraph{Training configuration.}
We did not perform a systematic hyperparameter search. One manually selected configuration was shared across folds, seeds, samplers, and query representations. For matched fold--seed--sampler runs, the flat and structured conditions differed only in their query representation.

\begin{itemize}
      \item \textbf{Initialization:} a shared-weight dual encoder initialized from ModernBERT-base. Query embeddings use the hidden state at the single \texttt{[MASK]} token; passage embeddings use mean pooling over attended positions after excluding sequence position zero. Both are $L^2$-normalized.

      \item \textbf{Optimization:} AdamW with learning rate $10^{-5}$, weight decay $0.01$, $\beta_1=0.9$, $\beta_2=0.999$, and $\epsilon=10^{-8}$. We used a linear learning-rate schedule, a warmup ratio of $0.1$, and maximum gradient norm $1.0$.

      \item \textbf{Schedule:} 20 epochs and three optimizer updates per epoch,
      for 60 updates in total.

      \item \textbf{Temperature:} $\tau=0.07$.

      \item \textbf{Positive sampling:} all gold passages are selected when a
      query has at most four; otherwise, four are selected without replacement.
      Selection is deterministic given the experiment seed, epoch, query
      identifier, and passage identifier, and is matched across query views and
      negative samplers.

      \item \textbf{Case-focused sampler:} 40 unique same-case negatives and
      20 unique other-case negatives. The other-case sample is passage-uniform
      over the eligible passages from all other training cases.

      \item \textbf{Pool-uniform sampler:} 60 unique negatives sampled passage-uniformly from all eligible passages in the training folds, including passages from the query's own case.

      \item \textbf{Sampling exclusions:} both samplers exclude every gold passage of the current query and sample without replacement. Each query therefore proposes 61--64 candidates: one to four selected positives and 60 negatives.

      \item \textbf{Cross-validation and seeds:} five case-disjoint rotations, with three folds used for training, one for validation, and one for testing. Each training rotation contains 294 queries from 24--26 cases. We used seeds 17, 29, and 43, two query representations, and two negative samplers, resulting in $5\times3\times2\times2=60$ controlled runs.

      \item \textbf{Checkpoint selection:} all 20 epochs were trained and evaluated. Checkpoints were selected lexicographically by (i) highest validation case-macro set Recall@20, (ii) highest validation case-macro full-ranking reciprocal rank of the first gold passage, and (iii) earliest epoch in the event of a tie.
  \end{itemize}

\clearpage

\section{Retrieval expanded results}
\label{sec:retrieval_expanded_results}
\begin{table}[!ht]
\centering
\small
\renewcommand{\arraystretch}{1.2}
\setlength{\tabcolsep}{3pt}
\begin{tabular*}{\columnwidth}{@{\extracolsep{\fill}}p{0.33\columnwidth}cccc@{}}
\toprule
\textbf{Candidate Pool} &
\textbf{\shortstack[c]{Hit rate\\@20}} &
\textbf{\shortstack[c]{Recall\\@20}} &
\textbf{\shortstack[c]{Complete\\recovery\\@20}} &
\textbf{\shortstack[c]{MRR}} \\
\midrule
\multicolumn{5}{@{}l}{\textbf{Same-case filtered}} \\
BM25 & 73.92 & 51.39 & 32.49 & 27.72 \\
E5-base-v2 & 68.74 & 46.27 & 29.00 & 19.93 \\
ModernBERT-base & 49.19 & 29.93 & 15.66 & 11.83 \\
Flat (case-focused) & \textbf{80.45} & \textbf{57.17} & \textbf{36.04} & 28.98 \\
Flat (pool-uniform) & 79.94 & 57.01 & 36.03 & 28.11 \\
Structured\\(case-focused) & 78.43 & 53.81 & 33.17 & \textbf{29.39} \\
Structured\\(pool-uniform) & 78.50 & 54.23 & 33.30 & 29.27 \\
\midrule
\multicolumn{5}{@{}l}{\textbf{Same-case full}} \\
BM25 & 55.21 & 33.33 & 18.67 & 9.35 \\
E5-base-v2 & 56.41 & 34.53 & 20.00 & 10.78 \\
ModernBERT-base & 40.15 & 22.03 & 10.36 & 9.46 \\
Flat (case-focused) & \textbf{67.98} & \textbf{43.98} & \textbf{26.07} & 18.68 \\
Flat (pool-uniform) & 66.98 & 43.20 & 25.34 & 17.93 \\
Structured\\(case-focused) & 62.70 & 39.60 & 22.17 & \textbf{18.87} \\
Structured\\(pool-uniform) & 63.00 & 39.68 & 22.21 & 18.47 \\
\midrule
\multicolumn{5}{@{}l}{\textbf{Fold-global}} \\
BM25 & \textbf{42.49} & \textbf{23.99} & \textbf{12.82} & \textbf{7.84} \\
E5-base-v2 & 35.65 & 20.04 & 11.05 & 7.64 \\
ModernBERT-base & 4.68 & 1.60 & 0.46 & 1.34 \\
Flat (case-focused) & 15.67 & 8.25 & 4.63 & 3.91 \\
Flat (pool-uniform) & 15.58 & 8.38 & 4.74 & 3.84 \\
Structured\\(case-focused) & 13.00 & 6.64 & 2.94 & 4.03 \\
Structured\\(pool-uniform) & 13.22 & 6.72 & 2.88 & 4.00 \\
\bottomrule
\end{tabular*}
\caption{Five-fold case-macro retrieval results are reported as percentages. Queries are averaged within held-out cases. For fine-tuned models, results are further averaged across seeds 17, 29, and 43 within each case before averaging across all 42 cases. Flat and Structured refer to the representations used to fine-tune ModernBERT, while \textit{case-focused} and \textit{pool-uniform} denote the positive and negative sample strategies used during training. Recall@20 represents the mean fraction of each query’s gold passages recovered in the top 20. Complete recovery@20 indicates the proportion of queries for which all gold passages are recovered in the top 20. MRR is untruncated and calculated using the rank of the first gold passage in the complete candidate-pool ranking.}
\label{tab:retrieval_cv_metrics_at_20}
\end{table}

\newpage

\section{Case-disjoint fold sizes}
\label{sec:case_disjoint_fold}
\begin{table}[h]
\centering
\small
\setlength{\tabcolsep}{3pt}
\begin{tabularx}{\columnwidth}{@{}l>{\centering\arraybackslash}X>{\centering\arraybackslash}X>{\centering\arraybackslash}X@{}}
\toprule
Evaluation fold & Training cases & Test cases & Test spans \\
\midrule
\multicolumn{4}{@{}l}{\textbf{5 classes}} \\
1 & 34 & 8 & 121 \\
2 & 34 & 8 & 161 \\
3 & 34 & 8 & 131 \\
4 & 33 & 9 & 140 \\
5 & 33 & 9 & 165 \\
\midrule
\multicolumn{4}{@{}l}{\textbf{4 classes}} \\
1 & 33 & 9 & 175 \\
2 & 33 & 9 & 168 \\
3 & 34 & 8 & 105 \\
4 & 35 & 7 & 129 \\
5 & 33 & 9 & 141 \\
\bottomrule
\end{tabularx}
\caption{Case-disjoint fold sizes for the classification experiments. The five- and four-class splits were generated independently using five-fold \texttt{StratifiedGroupKFold}. Training and test sets never overlap within any fold.}
\label{tab:classification_fold_summary}
\end{table}

\begin{table}[h]
\centering
\footnotesize
\setlength{\tabcolsep}{2.5pt}
\begin{tabularx}{\columnwidth}{
    @{}
    >{\raggedright\arraybackslash}X
    >{\centering\arraybackslash}X
    >{\centering\arraybackslash}X
    >{\centering\arraybackslash}X
    >{\centering\arraybackslash}X
    @{}
}
\toprule
Evaluation fold & Training cases & Validation cases & Test cases & Test queries \\
\midrule
1 & 24 & 9 & 9 & 98 \\
2 & 25 & 8 & 9 & 98 \\
3 & 26 & 8 & 8 & 98 \\
4 & 26 & 8 & 8 & 98 \\
5 & 25 & 9 & 8 & 98 \\
\bottomrule
\end{tabularx}
\caption{Case-disjoint fold sizes for the retrieval experiments. In each evaluation fold, one group of cases is used for testing, one for validation, and the remaining three groups are used for training. Each fold has 294 training, 98 validation, and 98 test queries.}
\label{tab:retrieval_fold_summary}
\end{table}

\section{Retrieval Uncertainty Analysis}
\label{sec:retrieval_uncertainty}
\begin{table}[!h]
\centering
\footnotesize
\setlength{\tabcolsep}{4pt}
\begin{tabular*}{\columnwidth}{@{\extracolsep{\fill}}lrr@{}}
\toprule
\multicolumn{3}{@{}l}{\textbf{Negative sampler: pool-uniform $-$ case-focused}} \\
Query representation & Estimate & 95\% interval \\
\midrule
Flat       & $-0.09$ & $[-1.47, 1.29]$ \\
Structured & $\phantom{-}0.23$ & $[-1.03, 1.62]$ \\
\midrule
\multicolumn{3}{@{}l}{\textbf{Query representation: Structured $-$ Flat}} \\
 Negative sampler & Estimate & 95\% interval \\
\midrule
Case-focused & $-2.68$ & $[-7.11, 1.62]$ \\
Pool-uniform & $-2.36$ & $[-6.08, 1.24]$ \\
\bottomrule
\end{tabular*}
\caption{Paired case-bootstrap contrasts for Fold-global Hit@20 using 10,000 resamples. Values are percentage points; all intervals include zero.}
\label{tab:retrieval_bootstrap_contrasts}
\end{table}

\clearpage

\twocolumn[{
\begin{minipage}{\textwidth}

\section{Path Agreement}
\label{sec:path_agreement}
\captionsetup{type=table}

\vspace{0.2in}
\centering
\footnotesize
\setlength{\tabcolsep}{3pt}
\renewcommand{\arraystretch}{1.05}
\begin{tabular*}{\textwidth}{@{\extracolsep{\fill}}lllrrrrr@{}}
\toprule
Path criterion & Pairing & Label match & $P_o$ & $P_+$ & $P_-$ & $\bar{P}_{\pm}$ & $\kappa$ \\
\midrule
Full transitive & 1:1 Edit & Blind & 0.85 & 0.40 & 0.91 & 0.66 & 0.31 \\
Full transitive & 1:1 Semantic & Blind & 0.86 & 0.46 & 0.92 & 0.69 & 0.38 \\
Full transitive & Relaxed $\geq 0.50$ & Blind & 0.90 & 0.63 & 0.94 & 0.79 & 0.57 \\
Full transitive & 1:1 Edit & Same label & 0.88 & 0.58 & 0.93 & 0.76 & 0.52 \\
Full transitive & 1:1 Semantic & Same label & 0.87 & 0.55 & 0.93 & 0.74 & 0.47 \\
Full transitive & Relaxed $\geq 0.50$ & Same label & 0.91 & 0.64 & 0.95 & 0.80 & 0.59 \\
Implicit bridge & 1:1 Edit & Blind & 0.90 & 0.10 & 0.95 & 0.52 & 0.05 \\
Implicit bridge & 1:1 Semantic & Blind & 0.91 & 0.16 & 0.95 & 0.55 & 0.11 \\
Implicit bridge & Relaxed $\geq 0.50$ & Blind & 0.91 & 0.38 & 0.95 & 0.67 & 0.34 \\
Implicit bridge & 1:1 Edit & Same label & 0.91 & 0.20 & 0.95 & 0.57 & 0.15 \\
Implicit bridge & 1:1 Semantic & Same label & 0.90 & 0.19 & 0.95 & 0.57 & 0.14 \\
Implicit bridge & Relaxed $\geq 0.50$ & Same label & 0.92 & 0.42 & 0.96 & 0.69 & 0.37 \\
\bottomrule
\end{tabular*}
\caption{Path agreement over all aligned node pairs. The full-transitive condition treats two aligned nodes as agreeing when each annotator makes the target reachable from the source, allowing paths of any length. The implicit-bridge condition counts only paths recovered through an implicit intermediate node. Path agreement is higher than direct-edge agreement because annotators often preserve reachability even when they choose different local edge decompositions. Relaxed pairing uses a minimum of 50\% overlap between passages to be paired.}
\label{tab:iaa_path_summary_revised}

\vspace{0.2in}

\section{Direct Edge Agreement}
\label{sec:direct_edge_agreement}
\captionsetup{type=table}

\vspace{0.2in}
\centering
\footnotesize
\setlength{\tabcolsep}{4pt}
\renewcommand{\arraystretch}{1.05}
\begin{tabular*}{\textwidth}{@{\extracolsep{\fill}}lllrrrrrr@{}}
\toprule
Pairing & Label match & Context type & Contexts & $P_o$ & $P_+$ & $P_-$ & $\bar{P}_{\pm}$ & $\kappa$ \\
\midrule
1:1 Edit & Blind & All pairs & 1120 & 0.91 & 0.06 & 0.96 & 0.51 & 0.02 \\
1:1 Edit & Blind & Edge union & 99 & 0.03 & 0.06 & -- & -- & -- \\
1:1 Semantic & Blind & All pairs & 1120 & 0.92 & 0.10 & 0.96 & 0.53 & 0.06 \\
1:1 Semantic & Blind & Edge union & 99 & 0.05 & 0.10 & -- & -- & -- \\
Relaxed $\geq 0.50$ & Blind & All pairs & 712 & 0.92 & 0.31 & 0.96 & 0.63 & 0.27 \\
Relaxed $\geq 0.50$ & Blind & Edge union & 66 & 0.18 & 0.31 & -- & -- & -- \\
1:1 Edit & Same label & All pairs & 1054 & 0.92 & 0.16 & 0.96 & 0.56 & 0.12 \\
1:1 Edit & Same label & Edge union & 94 & 0.09 & 0.16 & -- & -- & -- \\
1:1 Semantic & Same label & All pairs & 1054 & 0.92 & 0.15 & 0.96 & 0.55 & 0.12 \\
1:1 Semantic & Same label & Edge union & 96 & 0.08 & 0.15 & -- & -- & -- \\
Relaxed $\geq 0.50$ & Same label & All pairs & 576 & 0.93 & 0.39 & 0.97 & 0.68 & 0.36 \\
Relaxed $\geq 0.50$ & Same label & Edge union & 50 & 0.24 & 0.39 & -- & -- & -- \\
\bottomrule
\end{tabular*}
\caption{Direct-edge agreement on matched explicit-span contexts. \textit{Contexts} is the number of aligned source--target span pairs evaluated. We measure agreement as a binary decision over each context: whether a direct support edge exists from the source span to the target span. Edge-union rows evaluate only source--target pairs where at least one annotator proposed an edge; because $NN=0$ by construction, $P_-$, $\bar{P}_{\pm}$, and $\kappa$ are not reported for those rows. Relaxed pairing uses a minimum of 50\% overlap between passages to be paired.}
\label{tab:iaa_edge_summary}

\end{minipage}
}]

\clearpage

\twocolumn[{
\begin{minipage}{\textwidth}

\section{Structured Query and Example}
\label{sec:structured_query}
\captionsetup{type=figure}

\vspace{0.2in}

\centering
\setlength{\tabcolsep}{6pt}
\renewcommand{\arraystretch}{1.08}
\begin{tabularx}{\textwidth}{@{}>{\raggedright\arraybackslash}X >{\raggedright\arraybackslash}X@{}}
\textbf{Structured retrieval schema} & \textbf{Condensed example query} \\
\hline
\begin{minipage}[t]{\linewidth}
\raggedright\footnotesize
\sectag{[ARG]}\\
  \roletag{[ROOT]} \statustag{[MISSING]}\\
  \sectag{[TREE]}\\[0.35em]

  \texttt{[STEP]}\\
  \quad \roletag{[CONCL]} \roletag{[ANALYSIS]} \textit{analysis span 1}\\
  \quad \roletag{[PREMISE]} \roletag{[RULE]} \textit{rule span 1}\\
  \quad \roletag{[PREMISE]} \roletag{[ANALYSIS]} \textit{analysis span 2}\\
  \texttt{[/STEP]}\\[0.35em]

  \texttt{[STEP]}\\
  \quad \roletag{[CONCL]} \roletag{[ANALYSIS]} \textit{analysis span 3}\\
  \quad \roletag{[PREMISE]} \roletag{[ANALYSIS]} \textit{analysis span 4}\\
  \quad \roletag{[PREMISE]} \roletag{[RULE]} \textit{rule span 2}\\
  \quad \roletag{[PREMISE]} \roletag{[ANALYSIS]} \textit{analysis span 1}\\
  \texttt{[/STEP]}\\[0.35em]

  \texttt{[STEP]}\\
  \quad \roletag{[CONCL]} \roletag{[ANALYSIS]} \textit{analysis span 5}\\
  \quad \roletag{[PREMISE]} \roletag{[ANALYSIS]} \textit{analysis span 3}\\
  \quad \roletag{[PREMISE]} \roletag{[RULE]} \textit{rule span 3}\\
  \texttt{[/STEP]}\\[0.35em]

  \texttt{[STEP]}\\
  \quad \roletag{[CONCL]} \statustag{[MISSING]}\\
  \quad \roletag{[PREMISE]} \roletag{[ANALYSIS]} \textit{analysis span 5}\\
  \quad \roletag{[PREMISE]} \texttt{[IMPLICIT]} \roletag{[RULE]}\\
  \texttt{[/STEP]}\\
  \sectag{[/TREE]}\\[0.35em]

  \sectag{[FOCUS]}\\
  \texttt{[STEP]}\\
  \quad \roletag{[CONCL]} \masktag{[MASK]}\\
  \quad \roletag{[PREMISE]} \roletag{[ANALYSIS]} \textit{analysis span 5}\\
  \quad \roletag{[PREMISE]} \texttt{[IMPLICIT]} \roletag{[RULE]}\\
  \texttt{[/STEP]}\\
  \sectag{[/FOCUS]}\\
  \sectag{[/ARG]}

\end{minipage}
&
\begin{minipage}[t]{\linewidth}
\raggedright\footnotesize
\sectag{[ARG]}\\
  \roletag{[ROOT]} \statustag{[MISSING]}\\
  \sectag{[TREE]}\\[0.35em]

  \texttt{[STEP]}\\
  \quad \roletag{[CONCL]} \roletag{[ANALYSIS]} The only beneficial right that Ringwalt relinquished \ldots\\
  \quad \roletag{[PREMISE]} \roletag{[RULE]} I.R.C. \S\ 677(a)(2) provides that the grantor
  \ldots\\
  \quad \roletag{[PREMISE]} \roletag{[ANALYSIS]} In accordance with Ringwalt's powers as trustee, the 1967 liquidating distribution \ldots\\
  \texttt{[/STEP]}\\[0.35em]

  \texttt{[STEP]}\\
  \quad \roletag{[CONCL]} \roletag{[ANALYSIS]} The series of transactions that took place in the instant case appears governed by I.R.C. \S\ 368(a)(1)(D),\\
  \quad \roletag{[PREMISE]} \roletag{[ANALYSIS]} In determining that the continuity of interest \ldots adequately supported by the record.\\
  \quad \roletag{[PREMISE]} \roletag{[RULE]} Assessing continuity of interest \ldots Bondy v.~Commissioner, 269 F.2d 463, 466--67 (4th Cir.~1959);\\
  \quad \roletag{[PREMISE]} \roletag{[ANALYSIS]} The only beneficial right that Ringwalt relinquished \ldots\\
  \texttt{[/STEP]}\\[0.35em]

  \texttt{[STEP]}\\
  \quad \roletag{[CONCL]} \roletag{[ANALYSIS]} Under these circumstances, it appears that Ringwalt should be treated as the owner of the Clifford Trust \ldots\\
  \quad \roletag{[PREMISE]} \roletag{[ANALYSIS]} The series of transactions that took place in the instant case appears governed by I.R.C. \S\ 368(a)(1)(D),\\
  \quad \roletag{[PREMISE]} \roletag{[RULE]} I.R.C. \S\ 368(a)(1)(D), which defines a reorganization as \ldots\\
  \texttt{[/STEP]}\\[0.35em]

  \texttt{[STEP]}\\
  \quad \roletag{[CONCL]} \statustag{[MISSING]}\\
  \quad \roletag{[PREMISE]} \roletag{[ANALYSIS]} Under these circumstances, it appears
  \ldots defined by section 368, was satisfied.\\
  \quad \roletag{[PREMISE]} \texttt{[IMPLICIT]} \roletag{[RULE]}\\
  \texttt{[/STEP]}\\
  \sectag{[/TREE]}\\[0.35em]

  \sectag{[FOCUS]}\\
  \texttt{[STEP]}\\
  \quad \roletag{[CONCL]} \masktag{[MASK]}\\
  \quad \roletag{[PREMISE]} \roletag{[ANALYSIS]} Under these circumstances, it appears
  \ldots defined by section 368, was satisfied.\\
  \quad \roletag{[PREMISE]} \texttt{[IMPLICIT]} \roletag{[RULE]}\\
  \texttt{[/STEP]}\\
  \sectag{[/FOCUS]}\\
  \sectag{[/ARG]}
\end{minipage}
\end{tabularx}
\caption{Structured retrieval schema and condensed example. Bracketed strings are literal model-input tokens; italicized spans are variables, and \ldots\ marks omitted text. The unique \texttt{[MASK]} is the retrieval slot, whereas \texttt{[MISSING]} marks non-focus occurrences of the withheld node.}
\label{fig:struct_retrieval_linearized}

\end{minipage}
}]

\clearpage

\twocolumn[{
\begin{minipage}{\textwidth}

\section{Flat-masked query schema and example}
\label{sec:flat_query_structure}
\captionsetup{type=figure}

\vspace{0.2in}

\centering
\setlength{\tabcolsep}{6pt}
\renewcommand{\arraystretch}{1.08}
\begin{tabularx}{\textwidth}{@{}>{\raggedright\arraybackslash}X >{\raggedright\arraybackslash}X@{}}
\textbf{Flat-masked schema} & \textbf{Condensed example query} \\
\hline
\begin{minipage}[t]{\linewidth}
\raggedright\footnotesize
\sectag{argument}\\
  \roletag{root}: \statustag{missing}\\[0.35em]

  \sectag{context}\\
  \roletag{conclusion}: \roletag{analysis}: \textit{analysis span 1}\\
  \roletag{premise}: \roletag{rule}: \textit{rule span 1}\\
  \roletag{premise}: \roletag{analysis}: \textit{analysis span 2}\\[0.35em]

  \roletag{conclusion}: \roletag{analysis}: \textit{analysis span 3}\\
  \roletag{premise}: \roletag{analysis}: \textit{analysis span 4}\\
  \roletag{premise}: \roletag{rule}: \textit{rule span 2}\\
  \roletag{premise}: \roletag{analysis}: \textit{analysis span 1}\\[0.35em]

  \roletag{conclusion}: \roletag{analysis}: \textit{analysis span 5}\\
  \roletag{premise}: \roletag{analysis}: \textit{analysis span 3}\\
  \roletag{premise}: \roletag{rule}: \textit{rule span 3}\\[0.35em]

  \roletag{conclusion}: \statustag{missing}\\
  \roletag{premise}: \roletag{analysis}: \textit{analysis span 5}\\
  \roletag{premise}: implicit \roletag{rule}\\[0.35em]

  \sectag{focus}\\
  \roletag{conclusion}: \masktag{[MASK]}\\
  \roletag{premise}: \roletag{analysis}: \textit{analysis span 5}\\
  \roletag{premise}: implicit \roletag{rule}

\end{minipage}
&
\begin{minipage}[t]{\linewidth}
\raggedright\footnotesize
\sectag{argument}\\
\roletag{root}: \statustag{missing}\\[0.35em]

\sectag{context}\\
\roletag{conclusion}: \roletag{analysis}: The only beneficial right \ldots\\
\roletag{premise}: \roletag{rule}: I.R.C. \S\ 677(a)(2) provides that the grantor \ldots\\
\roletag{premise}: \roletag{analysis}: In accordance with Ringwalt's powers as trustee, the 1967 liquidating distribution received from dissolution of R \& L, Inc. was allocated to principal, held by the trust for future distribution and actually distributed to Ringwalt upon termination of the trust in 1969.\\[0.35em]

\roletag{conclusion}: \roletag{analysis}: The series of transactions that took place in the instant case appears governed by I.R.C. \S\ 368(a)(1)(D),\\
\roletag{premise}: \roletag{analysis}: In determining that the continuity of interest \ldots adequately supported by the record.\\
\roletag{premise}: \roletag{rule}: Assessing continuity of interest \ldots Bondy v.~Commissioner, 269 F.2d 463, 466--67 (4th Cir.~1959);\\
\roletag{premise}: \roletag{analysis}: The only beneficial right that \ldots entitled to total control after ten years.\\[0.35em]

\roletag{conclusion}: \roletag{analysis}: Under these circumstances, it \ldots by section 368, was satisfied.\\
\roletag{premise}: \roletag{analysis}: The series of transactions that took place in the instant case appears governed by I.R.C. \S\ 368(a)(1)(D),\\
\roletag{premise}: \roletag{rule}: I.R.C. \S\ 368(a)(1)(D), which defines a reorganization as: \ldots under section 354, 355, or 356\\[0.35em]

\roletag{conclusion}: \statustag{missing}\\
\roletag{premise}: \roletag{analysis}: Under these circumstances, it appears \ldots defined by section 368, was satisfied.\\
\roletag{premise}: implicit \roletag{rule}\\[0.35em]

\sectag{focus}\\
\roletag{conclusion}: \masktag{[MASK]}\\
\roletag{premise}: \roletag{analysis}: Under these circumstances, \ldots section 368, was satisfied.\\
\roletag{premise}: implicit \roletag{rule}
\end{minipage}
\end{tabularx}
\caption{Schematic structure and condensed example of the flat-masked query representation. The unique \texttt{[MASK]} marks the retrieval slot; \texttt{missing} marks non-focus occurrences of the withheld node.}
\label{fig:flat_retrieval_linearized}

\end{minipage}
}]

\clearpage

\clearpage

\twocolumn[{
\begin{minipage}{\textwidth}

\section{Cases Description}
\label{sec:data_statement}

\captionsetup{type=table}

\centering
\footnotesize
\setlength{\tabcolsep}{2.5pt}
\renewcommand{\arraystretch}{0.96}

\begin{tabular}{@{}
  >{\raggedright\arraybackslash}p{0.23\textwidth}
  >{\raggedright\arraybackslash}p{0.09\textwidth}
  c
  >{\raggedright\arraybackslash}p{0.28\textwidth}
  >{\raggedright\arraybackslash}p{0.30\textwidth}
@{}}
\toprule
Case & Court & Year & Key provisions & Main topics \\
\midrule
\emph{Helvering v. Minnesota Tea Co.}
& U.S. & 1935
& Revenue Act of 1928 \S~112(i)(1)(A)--(B)
& Substantially-all-assets transfer for stock and cash; material proprietary continuity; reorganization without dissolution. \\

\emph{Britt v. C.I.R.}
& 4th Cir. & 1940
& Revenue Act of 1926 \S~203(b)(3)--(4), 203(h)(1)(A); Revenue Act of 1932 \S~113(a)(6); Revenue Act of 1934 \S~113(a)(12)
& Substantially-all-assets reorganization; \S 203(b)(3) nonrecognition; redeemed-stock carryover basis; prior decision not preclusive. \\

\emph{Le Tulle v. Scofield}
& U.S. & 1940
& Revenue Act \S~112(b)(4), 112(g), 112(i)(1)(A)
& Asset transfer for cash and bonds; creditor rather than proprietary status; taxable sale rather than reorganization. \\

\emph{C.I.R. v. Segall}
& 6th Cir. & 1940
& Revenue Act of 1928 or 1932 \S~112(b)(4), 112(d)(1)
& Cash-and-debenture asset transfer; taxable sale rather than reorganization; creditor status regardless of debenture term; 1932 transaction timing. \\

\emph{National Rubber Machinery Co. v. U.S.}
& Ct. Cl. & 1941
& Revenue Act of 1928 \S\S~112(b)(4), 112(d)(1), 113(a)(6)--(8), 114
& Depreciation basis after asset acquisition; 80-percent control; binding stock-resale commitments; cost versus carryover basis. \\

\emph{Lyon, Inc. v. C.I.R.}
& 6th Cir. & 1942
& Revenue Act of 1928 \S\S~112(b)(5), 112(i)(1), 113(a)(6)--(7); Revenue Act of 1932 \S~113(a)(7); Revenue Act of 1934 \S~113(a)(12)
& Patent depreciation basis; carryover basis; business-purpose reorganization; separate tax-planning steps. \\

\emph{Glenn v. Courier-Journal Job Printing Co.}
& 6th Cir. & 1942
& Revenue Act of 1934 \S\S~23(f), 23(k), 112, 112(g)
& Stock-loss and bad-debt deductions; liquidation rather than reorganization; no continuity of ownership. \\

\emph{Cushman Motor Works v. C.I.R.}
& 8th Cir. & 1942
& Revenue Act of 1934 \S~112(g)(1)(C), 112(g)(1)(E), 112(g)(2), 112(h); Neb. Comp. Stat. \S\S~24-107, 24-220
& Dissolution and sheriff sale; no intercorporate transfer; ownership shift; no reorganization. \\

\emph{Roebling v. C.I.R.}
& 3d Cir. & 1944
& Revenue Act of 1938 \S~112(b)(3), 112(g)(1)(A); Treas. Reg. 101
& Stock-for-bonds statutory merger; continuity of interest; creditor rather than proprietary status; taxable gain. \\

\emph{Westfir Lumber Co. v. C.I.R.}
& Tax Ct. & 1946
& Revenue Act of 1936 \S~112(g)(1)(B); Revenue Act of 1939 \S~213(g)
& Substantially-all-assets acquisition for voting stock; nonassenting bondholder cash; integrated transfers; carryover basis. \\

\emph{Survaunt v. C.I.R.}
& 8th Cir. & 1947
& I.R.C. \S~112(g)(1)(D), 112(h); Reg. 103 \S\S~19.112(g)-1 and -2
& Reorganization rather than liquidation; integrated transaction; continuity and business purpose; carryover basis. \\

\emph{Lewis v. C.I.R.}
& 1st Cir. & 1947
& I.R.C. \S\S~112(b)(3), 112(c)(1)--(2), 112(g)(1)(D), 115(c), 117; Reg. 103 \S\S~19.112(g)-1 and -2
& Partial liquidation versus reorganization; business-purpose requirement; integrated transaction; remand for missing finding. \\

\emph{Nelson v. U.S.}
& Ct. Cl. & 1947
& Revenue Act of 1926 \S~203(b)(3)--(4), 203(c), 203(e)(1), 203(h)(1)(A)
& Substantially-all-assets reorganization; shareholder stock distribution under plan; later stock-sale basis; delayed liquidation distributions. \\

\emph{Lewis v. C.I.R.}
& 1st Cir. & 1949
& I.R.C. \S\S~112(b)(3), 112(c)(1)--(2), 112(g)(1)(D), 115(c), 117
& Type D reorganization; business-purpose doctrine; integrated transferor liquidation; boot and dividend treatment. \\

\emph{H. Grady Manning Trust v. C.I.R.}
& Tax Ct. & 1950
& I.R.C. \S\S~112(b)(3), 112(g)(1)(A), 112(g)(1)(D), 115(a), 115(g)
& Holding-company merger; stock-and-debenture exchange; business-purpose reorganization; debentures not dividends. \\

\bottomrule
\end{tabular}

\caption{Cases in the corpus. Key provisions are selected for material relevance.}
\label{tab:corp-reorg-cases}

\end{minipage}
}]

%-----------------------------

\twocolumn[{
\begin{minipage}{\textwidth}

\captionsetup{type=table}
\ContinuedFloat

\centering
\footnotesize
\setlength{\tabcolsep}{2.5pt}
\renewcommand{\arraystretch}{0.96}

\begin{tabular}{@{}
  >{\raggedright\arraybackslash}p{0.23\textwidth}
  >{\raggedright\arraybackslash}p{0.09\textwidth}
  c
  >{\raggedright\arraybackslash}p{0.28\textwidth}
  >{\raggedright\arraybackslash}p{0.30\textwidth}
@{}}
\toprule
Case & Court & Year & Key provisions & Main topics \\
\midrule

\emph{U.S. v. Arcade Co.}
& 6th Cir. & 1953
& I.R.C. \S~112(b)(4), 112(g)(1)(D), 112(g)(2); Reg. 103 \S~19.112(g)-6
& Dissolution and trustee transfer; no intercorporate transfer; invalid reorganization plan; professional-fee deduction. \\

\emph{Forest Hotel Corp. v. Fly}
& S.D. Miss. & 1953
& I.R.C. \S\S~23(a)(1), 111, 112(b)(3), 112(g)(1)(B); Reg. 101 art. 112(g)(1); Reg. 45 art. 109
& Integrated tax-free merger; continuity and solely-voting-stock test; leasehold amortization; estoppel rejected. \\

\emph{Becher v. C.I.R.}
& Tax Ct. & 1954
& I.R.C. \S\S~112(b)(3), 112(c)(1)--(2), 112(g)(1)(D), 115(c), 115(g), 115(i), 117
& Business-purpose reorganization; tax-free stock exchange; separate partial liquidation; dividend-equivalent cash distribution. \\

\emph{Pebble Springs Distilling Co. v. C.I.R.}
& 7th Cir. & 1956
& I.R.C. \S~112(b)(3)--(4), 112(c), 112(e), 112(g)(1)(D), 112(h)
& Controlled asset transfer; 80-percent control; Type D reorganization; claimed-loss nonrecognition. \\

\emph{National Bank of Commerce of Norfolk v. U.S.}
& E.D. Va. & 1958
& I.R.C. \S~113(a)(7) (1939); Revenue Act of 1926 \S~203(b)(3), 203(h)(1); Reg. 130 \S~40.458-2(b)
& Bank asset purchase; no substantially-all-assets reorganization; deposit goodwill basis; excess-profits invested capital. \\

\emph{Grede Foundries, Inc. v. U.S.}
& E.D. Wis. & 1962
& I.R.C. \S\S~311(a), 331(a)(1), 361(a), 368(a)(1)(C); Treas. Reg. \S\S~1.311-1, 1.368-1 and -2
& Integrated Liberty asset transfer and liquidation; Type C solely-voting-stock failure; refund denied. \\

\emph{Grubbs v. C.I.R.}
& Tax Ct. & 1962
& I.R.C. \S\S~301, 302(b)(2), 316, 354, 356(a)(2), 368(a)(1)(D), 368(c)
& Integrated Type D reorganization; stock-plus-cash exchange; dividend-equivalent distribution; control under \S 368(c). \\

\emph{Mills v. C.I.R.}
& 5th Cir. & 1964
& I.R.C. \S~368(a)(1)(B); former \S~112(g)(1)(B)
& Type B stock acquisition; fractional-share cash; solely-voting-stock requirement; substance over form. \\

\emph{Davant v. C.I.R.}
& 5th Cir. & 1966
& I.R.C. \S\S~301, 316, 331, 337, 354, 356(a)(2), 368(a)(1)(D), 368(a)(1)(F), 482
& Stock sale through conduit; integrated liquidation-reincorporation; Type D and F reorganizations; boot and dividend treatment. \\

\emph{Sharp v. U.S.}
& S.D. Tex. & 1966
& I.R.C. \S\S~331(a)(1), 356, 368(a)(1)(D), 368(c)
& Complete liquidation versus Type D reorganization; plan and substantially-all-assets requirements; capital gain versus dividend; bond-gain dispute. \\

\emph{Holliman v. U.S.}
& S.D. Ala. & 1967
& I.R.C. \S\S~368(a)(1)(F), 381(a)
& Type F reorganization; enterprise continuity; reduced unsecured debt; loss-carryback refund. \\

\emph{Stauffer's Estate v. C.I.R.}
& 9th Cir. & 1968
& I.R.C. \S\S~172(b), 368(a)(1)(F), 381(b)(3), 381(c)(1)(A); Treas. Reg. \S~1.381(b)-1(a)(2)
& Multi-corporation merger; Type F reorganization; postmerger NOL carryback; allocation to transferor operations. \\

\emph{King Enterprises, Inc. v. U.S.}
& Ct. Cl. & 1969
& I.R.C. \S\S~243(a)(1), 354(a)(1), 356(a)(1)--(2), 368(a)(1)(A)
& Stock acquisition and later merger as Type A reorganization; step transaction; boot as dividend; dividends-received deduction. \\

\emph{Calcote v. U.S.}
& D.N.J. & 1971
& I.R.C. \S\S~354(a)(1), 368(a)(1)(B), 368(b)
& Pre-1964 triangular Type B exchange; integrated stock transfers; party-to-reorganization requirement; continuity of interest. \\

\emph{Yoc Heating Corp. v. C.I.R.}
& Tax Ct. & 1973
& I.R.C. \S\S~332(b), 334(b)(2), 334(b)(4), 362(b), 368(a)(1)(D), 368(a)(1)(F), 368(c), 381(b)(3)
& Integrated stock purchase and asset transfer; Type D and F failures; stepped-up basis; NOL carryback denial. \\

\bottomrule
\end{tabular}

\caption[]{Cases in the corpus (continued).}
\label{tab:corp-reorg-cases-2}

\end{minipage}
}]

%-----------------------------

\twocolumn[{
\begin{minipage}{\textwidth}

\captionsetup{type=table}
\ContinuedFloat

\centering
\footnotesize
\setlength{\tabcolsep}{2.5pt}
\renewcommand{\arraystretch}{0.96}

\begin{tabular}{@{}
  >{\raggedright\arraybackslash}p{0.23\textwidth}
  >{\raggedright\arraybackslash}p{0.09\textwidth}
  c
  >{\raggedright\arraybackslash}p{0.28\textwidth}
  >{\raggedright\arraybackslash}p{0.30\textwidth}
@{}}
\toprule
Case & Court & Year & Key provisions & Main topics \\
\midrule

\emph{Swanson v. U.S.}
& 9th Cir. & 1973
& I.R.C. \S\S~331, 337, 354(a)(1), 354(b)(1)(A), 368(a)(1)(D)
& Old Stockton liquidation and New Stockton formation; no reorganization plan; substantially-all-assets failure; \S\S 331 and 337 treatment. \\

\emph{Performance Systems, Inc. v. U.S.}
& M.D. Tenn. & 1973
& I.R.C. \S\S~172, 332, 368(a)(1)(A), 368(a)(1)(F), 381(b); Treas. Reg. \S~1.381(b)-1(a)(2)
& Upstream parent-subsidiary merger; concurrent Type A, Type F, and \S 332 treatment; \S 381(b) NOL carryback. \\

\emph{West Side Federal Sav. \& Loan Ass'n of Fairview Park v. U.S.}
& 6th Cir. & 1974
& I.R.C. \S\S~368(a)(1)(A), 7701; Treas. Reg. \S~1.368-1(b)
& Savings-and-loan statutory merger; savings accounts as proprietary interests; continuity of interest; tax-free Type A reorganization. \\

\emph{Movielab, Inc. v. U.S.}
& Ct. Cl. & 1974
& I.R.C. \S\S~172(b), 332, 361, 368(a)(1)(F), 381(a), 381(b)(3)
& Parent-subsidiary statutory merger; concurrent Type F and \S 332 treatment; postmerger NOL carryback to subsidiary. \\

\emph{Stanton v. U.S.}
& 3d Cir. & 1975
& I.R.C. \S\S~1361(l)--(m), 331, 356, 368(a)(1)(D), 368(c)
& Subchapter R termination; liquidation-reincorporation; control despite spouse stock; retained-property boot. \\

\emph{Aetna Cas. \& Sur. Co. v. U.S.}
& 2d Cir. & 1976
& I.R.C. \S\S~172, 381(b)(3), 368(a)(1)(F)
& Shell-subsidiary merger; Type F reorganization; postreorganization NOL carryback; minority-shareholder redemption. \\

\emph{Ringwalt v. U.S.}
& 8th Cir. & 1977
& I.R.C. \S\S~331, 337, 354--356, 368(a)(1)(D), 677(a)(2)
& Liquidation-reincorporation; Type D reorganization; continuity and common control; grantor-trust ownership. \\

\emph{Atlas Tool Co., Inc. v. C.I.R.}
& 3d Cir. & 1980
& I.R.C. \S\S~331(a)(1), 337, 354, 356(a)(2), 368(a)(1)(D), 531--533, 6901(a)(1)(A); Treas. Reg. \S~1.368-2(g)
& Type D reorganization; boot dividend limit; accumulated-earnings tax; transferee liability. \\

\emph{General Housewares Corp. v. U.S.}
& 5th Cir. & 1980
& I.R.C. \S\S~301(c), 316(a), 331(a)(1), 337, 354, 356(a), 358(a)(1), 361(a), 368(a)(1)(C)
& Concurrent Type C reorganization and \S 337 liquidation; shareholder stock nonrecognition; cash boot as dividend. \\

\emph{Simon v. C.I.R.}
& 5th Cir. & 1981
& I.R.C. \S\S~331, 354, 354(b)(1), 356(a)(2), 368(a)(1)(D)
& Integrated asset and franchise transfer and dissolution; Type D reorganization; substantially-all-assets and plan requirements; distribution character. \\

\emph{Rose v. U.S.}
& 9th Cir. & 1981
& I.R.C. \S\S~331, 337, 354(a)(1), 354(b)(1), 356(a)(2), 368(a)(1)(D), 368(c)
& Type D reorganization despite liquidation; no tax-avoidance motive required; cash boot as dividend; summary judgment. \\

\emph{Russell v. C.I.R.}
& 6th Cir. & 1987
& I.R.C. \S\S~332, 334(b)(2), 351(a), 354(a)(1), 368(a)(1)(B), 368(a)(1)(F), 368(c), 381(b)(3); Treas. Reg. \S~1.368-2(a)
& \S\S 332 and 334(b)(2) liquidation; stepped-up basis; step transaction; NOL carryback and \S 351 denial. \\
\bottomrule
\end{tabular}

\caption[]{Cases in the corpus (continued).}
\label{tab:corp-reorg-cases-3}

\end{minipage}
}]

\clearpage

\twocolumn[{
\begin{minipage}{\textwidth}

\section{Related Work Comparison}
\label{sec:related_work}

\captionsetup{type=table}

\centering
\small
\setlength{\tabcolsep}{2pt}
\renewcommand{\arraystretch}{1.08}

\begin{tabularx}{\textwidth}{@{}
  >{\raggedright\arraybackslash}p{0.13\textwidth}
  >{\raggedright\arraybackslash}p{0.12\textwidth}
  >{\raggedright\arraybackslash}p{0.17\textwidth}
  >{\raggedright\arraybackslash}p{0.13\textwidth}
  >{\raggedright\arraybackslash}p{0.13\textwidth}
  >{\raggedright\arraybackslash}X
@{}}

\toprule
\textbf{Work} &
\textbf{Domain} &
\textbf{Schema and granularity} &
\textbf{Goal} &
\textbf{Structure layer} &
\textbf{How our paper differs} \\
\midrule

\textbf{This paper} & U.S. federal tax opinions on corporate reorganizations under I.R.C.~\S~368 & Free spans plus sentence, flat, and tree views; \textit{Rule}, \textit{Analysis}, \textit{Conclusion}, \textit{Background Facts}, \textit{Procedural History} & Corpus release, functional-role classification, and argument-completion retrieval & Directed support trees with implicit intermediate conclusions; reliability separately reported for labels and structure & Narrow but underrepresented U.S. tax-law domain; expert-adjudicated support trees; explicit reliability limits for structural annotations; case-disjoint classification and retrieval experiments. \\
\midrule
\citet{lawrence-reed-2019-argument} & General argument mining & Survey-level taxonomy, not a legal annotation corpus & Defines and surveys argument mining & Not a corpus-specific structural layer & Provides background framing; our paper operationalizes argument mining in a concrete U.S. tax-law corpus with annotated roles and support trees. \\
\citet{yamada-etal-2019-corpus} & Japanese judgment documents & Legal argumentation corpus for structure-based summarization & Structure-based legal summarization & Argument structure for summarization & Different jurisdiction and downstream goal; our paper targets U.S. federal tax opinions and evaluates both functional-role learning and argument-completion retrieval. \\
\citet{poudyal-etal-2020-echr} & European Court of Human Rights decisions and judgments & Clause-level premise, conclusion, and non-argument labels & Legal argument-mining corpus & Premise--conclusion argument groupings; no typed support or attack labels & ECHR human-rights domain and clause-level argumentative labels; our paper uses functional legal-role spans and sentences in U.S. tax opinions and organizes argumentative spans into support trees. \\
\citet{grundler-etal-2022-detecting} & CJEU fiscal state-aid decisions & Argumentative elements, element types, and argument schemes & Argument detection in EU fiscal state-aid law & Argument-scheme labels on legal premises; no explicit support or attack edges & Fiscal-law subject matter, but EU state-aid rather than U.S. corporate-reorganization tax law; our paper uses a smaller functional-role inventory and support-tree representation. \\
\citet{santin-2023-relation-prediction} & CJEU fiscal state-aid decisions & Argument-structure representation extending the Demosthenes setting & Directed link prediction between known argumentative components & Five typed inferential-link relations among sentence-level components & Focuses on relation prediction in CJEU state-aid decisions; our paper contributes a U.S. tax corpus and evaluates within-case and split-global argument-completion retrieval from support trees. \\
\citet{habernal-etal-2024-mining} & European Court of Human Rights decisions & Span-level actor--argument-type scheme with 16 legally grounded argument types & Mining legal arguments in court decisions & Flat argument-type and actor labels; no relation graph or explicit support or attack edges & Rich ECHR-specific argument-type taxonomy; our paper instead uses compact functional legal roles and studies support-tree reliability in U.S. tax case law. \\
\citet{chen-etal-2026-guidelines} (arXiv preprint) & Chinese judicial decisions & Proposition types plus relations such as support, attack, joint, match, and identity & Annotation and visualization guidelines & Multi-relation argumentation structures & Broader relation taxonomy and Chinese-law setting; our paper constrains relations to directed support trees for U.S. tax opinions and reports empirical agreement and downstream experiments. \\

\bottomrule
\end{tabularx}

\caption{Comparison with related work. The table emphasizes differences in domain, annotation schema, structural representation, and evaluation setup.}
\label{tab:related_work_comparison}

\end{minipage}
}]

\twocolumn[{
\begin{minipage}{\textwidth}

\captionsetup{type=table}
\ContinuedFloat

\centering
\small
\setlength{\tabcolsep}{2pt}
\renewcommand{\arraystretch}{1.08}

\begin{tabularx}{\textwidth}{@{}
  >{\raggedright\arraybackslash}p{0.13\textwidth}
  >{\raggedright\arraybackslash}p{0.12\textwidth}
  >{\raggedright\arraybackslash}p{0.17\textwidth}
  >{\raggedright\arraybackslash}p{0.13\textwidth}
  >{\raggedright\arraybackslash}p{0.13\textwidth}
  >{\raggedright\arraybackslash}X
@{}}

\toprule
\textbf{Work} &
\textbf{Domain} &
\textbf{Schema and granularity} &
\textbf{Goal} &
\textbf{Structure layer} &
\textbf{How our paper differs} \\
\midrule

\citet{walker-etal-2017-semantic} & U.S. veterans' claims decisions & Sentence roles and propositional connective types & Computational legal reasoning resource & Connective propositional relations & U.S. legal setting, but different administrative domain and sentence-role and connective focus; our paper targets corporate tax opinions with span-level support trees and sentence-level derived views. \\

\citet{xu-etal-2020-using} & Canadian case decisions and expert-written summaries & Sentence-level \textit{Issue}, \textit{Reason}, \textit{Conclusion}, and non-IRC labels & Sentence-role classification toward extractive summarization & No explicit links among Issue--Reason--Conclusion sentences & Uses role classification as a step toward summarization; our paper releases a domain-specific corpus and tests classification plus argument-completion retrieval. \\

\citet{xu-ashley-2022-multigranularity} & Canadian case decisions and expert-written summaries & Sentence-level Issue--Reason--Conclusion labels with BIO token labels derived from them & Token- and sentence-level rhetorical-role classification & Compares token and sentence representations; no explicit support or IRC-link layer & Emphasizes granularity for legal AM; our paper provides adjudicated U.S. tax annotations in multiple released views and explicitly separates reliable node labels from lower-confidence edges. \\
\citet{bhattacharya-etal-2019-identification} & Indian Supreme Court judgments & Sentence-level rhetorical roles & Rhetorical-role identification & No explicit support-link layer & Rhetorical segmentation without argument-support trees; our paper links rules and analyses to conclusions in U.S. federal tax opinions. \\
\citet{kalamkar-etal-2022-corpus} & Indian legal documents & Rhetorical-role and document-structure labels & Automatic legal-document structuring & No explicit support-relation layer & Multi-court rhetorical-role resource; our paper focuses on functional legal roles plus support-tree structure in a narrow U.S. tax domain. \\
\citet{malik-etal-2022-semantic} & Indian legal documents & Rhetorical-role semantic segmentation & Semantic segmentation of legal documents & No explicit support-link layer & Labels discourse function but not premise-to-conclusion support; our paper annotates both role labels and directed support among argumentative spans. \\
\citet{nigam-etal-2025-legalseg} & Indian legal judgments & Rhetorical-role classification for judgment structure & Segmenting legal judgments & No explicit support-link layer & Focuses on rhetorical-role classification at scale; our paper adds explicit support-tree annotations and retrieval tasks tied to missing argumentative supports. \\
\citet{savelka-ashley-2018-segmenting} & U.S. trade-secret and cyber-crime opinions & Consecutive, non-overlapping segments with seven functional and issue-specific labels & CRF-based segmentation into functional parts & No arbitrary span or support-relation annotation & U.S. case-law precedent, but different substantive domains and segmentation design; our paper permits arbitrary spans and directed support trees in corporate tax opinions. \\
\citet{csanyi-etal-2025-rrl} & Hungarian judicial decisions & Sentence-level rhetorical-role labels & Rhetorical-role classification & No explicit support-link layer & Different jurisdiction and no structural argument layer; our paper targets U.S. tax-law reasoning with functional roles and tree-structured support. \\
\citet{bambroo-etal-2025-marro} & Indian and U.K. Supreme Court datasets & Sentence-level rhetorical-role classification & Multi-task model for rhetorical roles & No explicit support-relation layer & A modeling contribution that also adds 100 newly annotated Indian judgments; our paper contributes a new U.S. tax corpus and evaluates whether annotated structure helps retrieval. \\

\bottomrule
\end{tabularx}

\caption[]{Comparison with related work (continued).}
\label{tab:related_work_comparison_2}

\end{minipage}
}]

%-------------------------------

\twocolumn[{
\begin{minipage}{\textwidth}

\captionsetup{type=table}
\ContinuedFloat

\centering
\small
\setlength{\tabcolsep}{2pt}
\renewcommand{\arraystretch}{1.08}

\begin{tabularx}{\textwidth}{@{}
  >{\raggedright\arraybackslash}p{0.13\textwidth}
  >{\raggedright\arraybackslash}p{0.12\textwidth}
  >{\raggedright\arraybackslash}p{0.17\textwidth}
  >{\raggedright\arraybackslash}p{0.13\textwidth}
  >{\raggedright\arraybackslash}p{0.13\textwidth}
  >{\raggedright\arraybackslash}X
@{}}

\toprule
\textbf{Work} &
\textbf{Domain} &
\textbf{Schema and granularity} &
\textbf{Goal} &
\textbf{Structure layer} &
\textbf{How our paper differs} \\
\midrule

\citet{belfathi-etal-2026-coupling} & U.S. Supreme Court opinions & Three-level rhetorical-role granularity; sentences labeled by rhetorical function & Hierarchical rhetorical-role labeling & Hierarchical label taxonomy and sentence sequence; no explicit support-relation layer & U.S. and multi-granular, but broad U.S. Supreme Court rhetorical roles; our paper is domain-specific to federal tax reorganizations and includes support-tree annotations. \\
\citet{guha-etal-2023-legalbench} & Broad legal-reasoning benchmark & Task-based benchmark built with legal professionals & Measuring legal reasoning in LLMs & Benchmark tasks, not document-level support trees & Broad evaluation suite rather than an annotated case-law corpus; our paper provides case-level annotations and domain-specific retrieval and classification tasks. \\

\citet{Kang2024BridgingLA} & Malaysian contract-law scenario analysis & Semi-structured Issue--Rule--Application--Conclusion (IRAC) methodology & Augmenting reasoning with IRAC-style data & IRAC-stage annotations plus a semi-structured legal knowledge graph; not opinion-level support trees & Uses IRAC to structure legal reasoning tasks; our paper adapts functional roles to actual U.S. tax opinions and annotates support paths among spans. \\
\citet{jang-etal-2025-pilot} & U.S. PTAB ex parte patent appeals & Case-level Issue, Rule, and Conclusion classification tasks (Application omitted) & Legal-reasoning benchmark & Independent case-level labels; no explicit support-relation layer & Patent benchmark rather than tax case-law corpus; our paper studies corporate-reorganization opinions and support-tree-based argument completion. \\

\citet{choi-etal-2026-taxation} & Korean additional-tax-penalty questions & Case-level facts and claims rendered as binary, rationale-choice, and essay tasks; IRAC is used for analysis and rubrics & Evaluate LLM legal reasoning about penalty-exemption lawfulness & Parallel case-level task views; no document support-relation layer & Shares tax orientation, but focuses on Korean penalty-exemption questions; our paper annotates judicial opinions about corporate reorganizations under I.R.C.~\S~368. \\
\citet{yu-etal-2022-legal-prompting} (arXiv preprint) & Legal entailment and prompting & IRAC-derived prompts & Improving model reasoning through prompting & Prompt structure, not annotated document relations & Methodological prompting work; our paper contributes data annotations and evaluates learnability and retrieval from legal-role and support-tree representations. \\
\citet{servantez-etal-2024-chain} & Rule-based legal reasoning with LLMs & Chain of Logic, IRAC-inspired prompting & Prompting method for rule-based reasoning & Prompt-level reasoning chain & Focuses on inference prompting; our paper creates an annotated corpus of real opinions with explicit support links and reliability analysis. \\
\citet{bongard-etal-2022-legal} & U.S. civil procedure & Topic introductions, civil-procedure fact patterns, answer candidates, and explanatory analyses & Binary answer-correctness classification for legal educational problems & Task instances rather than document-level support trees & U.S. legal-reasoning task, but civil procedure and constructed task format; our paper annotates full tax opinions and derives retrieval queries from support trees. \\
\citet{hou-etal-2025-clerc} & U.S. federal case-law retrieval and retrieval-augmented generation (RAG) & Masked citation contexts linked to cited federal case documents and derived passages & Retrieve cited case authorities and generate citation-conditioned legal analysis & Directed citing-case-to-cited-authority edges; no proposition-level support labels & Retrieval-related U.S. case-law resource; our paper instead performs within-case and split-global passage retrieval for missing argument supports derived from annotated trees. \\

\bottomrule
\end{tabularx}

\caption[]{Comparison with related work (continued).}
\label{tab:related_work_comparison_3}

\end{minipage}
}]

\clearpage

\clearpage

\twocolumn[{
\begin{minipage}{\textwidth}

\section{Classification Experiment Prompt}
\label{sec:classification_prompt}

\begin{promptbox}[label={prompt:case-classifier-system}]
{Passage-classification prompt: system instruction and guidelines (part 1 of 2)}
# Passed separately as `instructions` in the Responses API call.
_SYSTEM_PROMPT = (
    "You are a precise legal passage classifier for U.S. judicial opinions. "
    "Choose exactly ONE label from the allowed set and output ONLY one line in the format "
    "Class:<label>. No extra words, no punctuation, no explanation."
)

_DEFAULT_ALLOWED_LABELS = (
    "Analysis",
    "Background Facts",
    "Conclusion",
    "Procedural History",
    "Rule",
)

# The four-class setting removes Conclusion from this inventory.
def _get_allowed_labels(combine_analysis_conclusion):
    labels = list(_DEFAULT_ALLOWED_LABELS)
    if combine_analysis_conclusion:
        labels = [label for label in labels if label != "Conclusion"]
    return labels

_GUIDELINES_CORE = (
    "Annotation scheme (match human annotators): spans are labeled by FUNCTIONAL ROLE in a "
    "chain of syllogisms (polysyllogistic IRAC). Rules and Analyses are argument nodes; "
    "Background Facts and Procedural History are contextual blocks and are not part of the reasoning chain.\n\n"
    "LABEL DEFINITIONS (use these meanings):\n"
    "- Background Facts: narrative context about what happened outside the courtroom (events, transactions, parties, dates). "
    "Includes IRS/agency administrative steps (audits, assessments, refund claims/denials). "
    "These spans inform the reader but do not themselves apply a rule or draw an inference toward the holding.\n"
    "- Procedural History: court/litigation process and posture (complaints, motions, hearings, judgments, appeals, remands, "
    "petitions for certiorari/grants). Focus is the procedural timeline in court. "
    "IRS administrative steps are NOT procedural history.\n"
    "- Rule: a generally applicable premise used to justify an inference\u2014statutes, regulations, precedent holdings, tests, "
    "definitions, and other reusable generalizations (including implicit/brute premises that license an inference). "
    "Summaries of precedent (facts/holdings) used as authority count as Rule. Citations often appear but are not required.\n"
    "- Analysis: case-specific reasoning that applies/interprets a Rule using this case\u2019s facts/record; evaluates evidence; "
    "accepts/rejects/distinguishes arguments; draws causal/logical inferences. "
    "IMPORTANT: intermediate/local conclusions in a reasoning chain are labeled Analysis (even if phrased 'we conclude...' "
    "or 'therefore...') when they support a later step.\n\n"
    "TIE-BREAKERS:\n"
    "1) Court procedure => Procedural History. IRS/admin steps => Background Facts.\n"
    "2) Stating a legal standard / definition / precedent holding => Rule. Applying or distinguishing it here => Analysis.\n"
    "3) If both appear, choose the dominant function: 'state the law/test' => Rule; 'apply to facts / infer' => Analysis.\n"
)

\end{promptbox}

\end{minipage}
}]

\clearpage

\twocolumn[{
\begin{minipage}{\textwidth}

\begin{promptbox}[label={prompt:case-classifier-renderer}]
{Passage-classification prompt: renderer and variants (part 2 of 2)}
_CONCLUSION_GUIDELINE = (
    "CONCLUSION (only when this label exists in the allowed set):\n"
    "- Conclusion: ONLY the terminal outcome of an argument tree / issue\u2014i.e., the court\u2019s ultimate holding or disposition "
    "(e.g., 'judgment affirmed/reversed', 'summary judgment granted', 'the deduction is allowed/denied'). ")
def _format_label_options(labels):
    if not labels:
        raise ValueError("At least one label must be provided.")
    if len(labels) == 1:
        return labels[0]
    return ", ".join(labels[:-1]) + f", or {labels[-1]}"
def _render_prompt(
    *,
    context,
    passage,
    context_mode,
    allowed_labels,
    include_conclusion_guidance,
):
    guidelines = _GUIDELINES_CORE
    if include_conclusion_guidance:
        guidelines += _CONCLUSION_GUIDELINE
    label_options = _format_label_options(list(allowed_labels))
    has_conclusion = any(lbl.lower() == "conclusion" for lbl in allowed_labels)
    conclusion_fallback = ""
    if not has_conclusion:
        conclusion_fallback = (
            "IMPORTANT: 'Conclusion' is NOT an available label in this run. "
            "If the passage states the final outcome/disposition, label it as Analysis.\n\n")

    task_intro = (
        "TASK\n"
        "You will be given one TARGET PASSAGE from a case.\n"
        "Use the label definitions to pick exactly one class.\n")
    if context_mode == "case_context":
        task_intro = (
            "TASK\n"
            "You will be given (1) case text (context) and (2) one TARGET PASSAGE from that case.\n")
    prompt = (
        task_intro
        + f"Choose exactly ONE label from: {label_options}\n\n"
        "OUTPUT FORMAT (STRICT)\n"
        "Return exactly ONE line:\n"
        "Class:<label>\n"
        "Do not output anything else.\n\n"
        f"{conclusion_fallback}"
        "GUIDELINES (match the human annotators)\n"
        f"{guidelines}\n\n")
    if context_mode == "case_context":
        prompt += (
            "<<<CASE_TEXT>>>\n"
            f"{context}\n"
            "<<<END_CASE_TEXT>>>\n\n")
    prompt += (
        "<<<TARGET_PASSAGE>>>\n"
        f"{passage}\n"
        "<<<END_TARGET_PASSAGE>>>\n")
    return prompt

\end{promptbox}

\end{minipage}
}]

\end{document}